\documentclass[10pt,twocolumn,letterpaper]{article}

\usepackage[pagenumbers]{cvpr} 

\usepackage{graphicx}
\usepackage{amsmath}
\usepackage{amssymb}
\usepackage{booktabs}

\usepackage{multirow}
\usepackage{diagbox} 
\usepackage[pagebackref,breaklinks,colorlinks]{hyperref}

\usepackage[capitalize]{cleveref}
\crefname{section}{Sec.}{Secs.}
\Crefname{section}{Section}{Sections}
\Crefname{table}{Table}{Tables}
\crefname{table}{Tab.}{Tabs.}

\usepackage[accsupp]{axessibility}  
\iftrue
\usepackage{color}

\definecolor{pink}{rgb}{0.858, 0.188, 0.478}

\newcommand{\BR}[1]{{\color{pink}{\bf Bog:} #1}}
\newcommand{\HT}[1]{{\color{red}{\bf Hec:} #1}}
\newcommand{\YX}[1]{{\color{blue}{\bf YX:} #1}}
\newcommand{\alertJW}[1]{{\color{magenta}{\bf JW:} #1}}
\newcommand{\alertLL}[1]{{\color{yellow}{\bf LL:} #1}}

\else
\newcommand{\alertJW}[1]{}
\newcommand{\YX}[1]{}
\newcommand{\BR}[1]{}
\newcommand{\HT}[1]{}
\newcommand{\alertLL}[1]{}
\fi

\newcommand{\minisection}[1]{\vspace{0.04in} \noindent {\bf #1}}
\newcommand*{\affmark}[1][*]{\textsuperscript{#1}}

\def\cvprPaperID{33} 
\def\confName{CVPR}
\def\confYear{2022}

\begin{document}

\title{Transferring Unconditional to Conditional GANs with Hyper-Modulation}
\author{Héctor Laria\affmark[1]\ \qquad Yaxing Wang\affmark[2] \qquad Joost van de Weijer\affmark[1] \qquad Bogdan Raducanu\affmark[1]\\
\affmark[1]~Computer Vision Center, Barcelona, Spain\\
\affmark[2]~Nankai University, China\\
{\tt\small \{hlaria, yaxing, joost, bogdan\}@cvc.uab.es}
}
\maketitle

\begin{abstract}
GANs have matured in recent years and are able to generate high-resolution, realistic images. However, the computational resources and the data required for the training of high-quality GANs are enormous, and the study of transfer learning of these models is therefore an urgent topic. 
Many of the available high-quality pretrained GANs are unconditional (like StyleGAN). For many applications, however, conditional GANs are preferable, because they provide more control over the generation process, despite often suffering more training difficulties.
Therefore, in this paper, we focus on transferring from high-quality pretrained unconditional GANs to conditional GANs. 
This requires architectural adaptation of the pretrained GAN to perform the conditioning. To this end, we propose hyper-modulated generative networks that allow for shared and complementary supervision.
To prevent the additional weights of the hypernetwork to overfit, with subsequent mode collapse on small target domains, we introduce a self-initialization procedure that does not require any real data to initialize the hypernetwork parameters. To further improve the sample efficiency of the transfer, we apply contrastive learning in the discriminator, which effectively works on very limited batch sizes. In extensive experiments, we validate the efficiency of the hypernetworks, self-initialization and contrastive loss for knowledge transfer on standard benchmarks.
Our code is available at \url{https://github.com/hecoding/Hyper-Modulation}.

\end{abstract}

\section{Introduction}

Generative Adversarial Networks (GANs) have become ubiquitous in a vast array of applications due to their modelling and synthesis power.
Current high-quality GANs consist of several millions of parameters \cite{karras2020analyzing}. In this magnitude range, the training of these models quickly become prohibitive in terms of computing resources and amount of training data required.
Transfer learning for generative models explores how the knowledge of pretrained GANs can be transferred to new domains, potentially with much fewer training samples. 

\begin{figure}
    \centering
      \includegraphics[width=\linewidth]{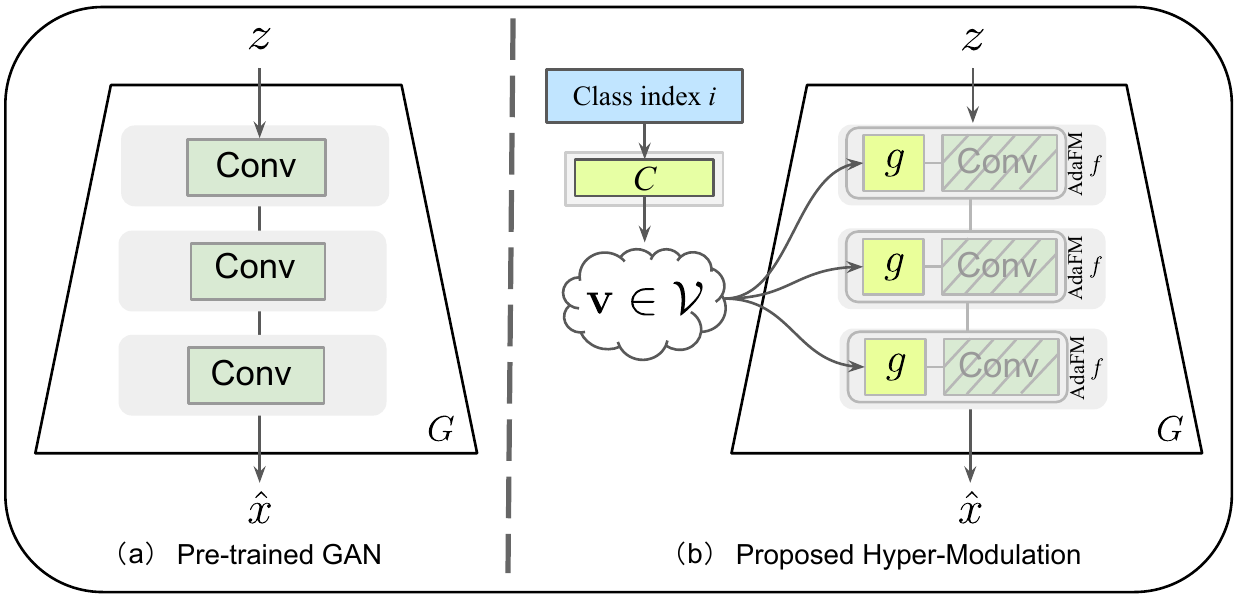}
    \caption{Transfer learning from unconditional (a) to conditional GAN (b), 
    via on-the-fly modulation of the pre-trained weights.
    The class network \textit{C} outputs a point $\mathbf{v}$ in the class space given a label $i$, which is then fed into the modulator \textit{g} for domain-specific generation. 
    }
    \label{fig:general}
\end{figure}

In the transfer learning area of generative models, Wang \etal~\cite{wang2018transferring} initially investigated unconditional transferring by finetuning a pre-trained GAN to a target domain. Further research improved the quality of transfer learning to small domains by reducing the number of learnable parameters~\cite{noguchi2019image, mo2020freeze, zhao2020leveraging} or by identifying the subspace of a pre-trained GAN that best models the target data~\cite{wang2020minegan}. The majority of efforts (see Table~\ref{tab:overview}) have been driven towards transferring knowledge from unconditional GANs to also unconditional GANs (single source and target), from conditional to unconditional \cite{wang2020minegan} (multiple sources, single target), which considers transferring a pre-trained cGAN to a single-class target domain, and from conditional to conditional \cite{Shahbazi_2021_CVPR} (multiple sources and targets), which proposes a method to transfer between conditional GANs through linear combination of conditionings.

In this work, we investigate the knowledge transfer from an unconditional to a conditional GAN. This setup is especially relevant, because of the availability of many high-quality unconditional pretrained GANs. There exist pretrained conditional models (cGANs), however, they have not seen adoption as widely as unconditional ones since they suffer from unstable performance among training runs~\cite{Brock19}, data and computational resources needed are higher, and do not employ an intermediate latent space, which is essential for GAN-based image editing \cite{perarnau2016invertible,shen2019interpreting,shu2017neural,zhu2020domain}. On the other hand, for many applications, it is required that the generation process should be conditional. Therefore, we investigate the transfer from unconditional pretrained GAN models to conditional GANs. An additional benefit of transferring to conditional GANs (when compared to transferring to multiple unconditional GANs) is the fact that they enable the sharing of weights between the multiple classes, thereby exploiting the similarities between the various classes.

In this paper, we leverage weight modulation from the context of continual learning \cite{Cong20,Perez18} to transform an unconditional source GAN to a cGAN, as depicted in \cref{fig:general}. 
Our method allows for efficient transfer learning, 
where frozen pre-trained weights are conditionally modulated to yield target-specific outputs. However, a drawback of this approach is that the class-specific modulation parameters are learned independent of each other. To exploit the existing similarities among the multiple classes of the target domain,
we propose the use of hypernetworks~\cite{HaDL17}. Hypernetworks have been proven efficient on diverse areas, from multi-task learning\cite{DBLP:journals/corr/abs-1809-10889,Meyerson19,Shen18} to continual learning \cite{Oswald20}, delivering additional improvements on weight pruning \cite{ZechunLiu19} over traditional networks. Yet to our knowledge, they have not been applied to transfer learning. In this work, we aim to show that hypernetworks can result in more efficient knowledge transfer to multi-class domains, due to their intrinsic knowledge sharing among layers \cite{HaDL17,Oswald20}. However, the hypernetwork introduces new parameters that need to be trained from scratch to even regain the source generation power. 
To initialize these parameters, we propose a \textit{self-alignment} method that learns well-initialized hypernetworks without getting access to any real data. Furthermore, we introduce contrastive learning in the discriminator for quality improvement like other generative methods propose \cite{yu2021dual,NEURIPS2020_f490c742,Jeong21}, except that this effectively works with very limited batch sizes, \ie 10 samples, contrary to current literature \cite{JMLR:v13:gutmann12a,Chen20,NEURIPS2020_f490c742}.

In summary, we propose the following contributions.
\begin{itemize}
    \item We are the first to investigate knowledge transfer from unconditional to conditional GANs. 
    \item We propose a new method based on hypernetworks and adaptive weight modulation that efficiently transfers unconditional to conditional GANs. 
    \item In addition, we propose an approach for self-initialization of the hypernetwork parameters, that further allows applying a contrastive loss to the GAN discriminator with tiny batch sizes. Both these novelties result in significant improvements of the knowledge transfer.
    \item Results on several datasets show that we outperform existing methods and that FID improves on several datasets (including a notable drop of 30 points on the AFHQ dataset).
\end{itemize}

\section{Related work}
\label{sec:related-work}

\begin{table}
  \setlength{\tabcolsep}{20pt}
  \centering
  \resizebox{\columnwidth}{!}{
  \begin{tabular}{@{}cccl@{}}
    \toprule
    Method & Source & Target \\
    \midrule
    TransferGAN \cite{wang2018transferring}, MineGAN\cite{wang2020minegan}& U & U \\
 AdaFM~\cite{zhao2020leveraging},FreezeD~\cite{mo2020freeze}, BSA~\cite{noguchi2019image}&  &      \\ EWCGAN~\cite{li2020few}, CDCGAN~\cite{ojha2021few}&  &      \\
 \midrule
    MineGAN \cite{wang2020minegan} & C & U  \\
    cGANTransfer \cite{Shahbazi_2021_CVPR} & C & C \\
    Hyper-Modulation (Ours) & U & C  \\
    \bottomrule
  \end{tabular}
  }
  \caption{Overview of existing transfer learning methods for GANs according to  whether involved GANs for source and target domain are unconditional (U) or conditional (C). Even though transfer learning for GANs has seen an increased research activity, transferring unconditional to conditional has not been addressed before. The existence of high-quality unsupervised models \cite{karras2020analyzing}
  -- that are the state of the art in high-resolution image generation -- makes their transfer to conditional target domains especially pertinent.}
  \label{tab:overview}
\end{table}

\minisection{Generative adversarial networks.} GANs play a minimax game~\cite{goodfellow2014generative} between a generator and discriminator. The discriminator aims to tell the real distribution and the fake one apart, while the generator tries to synthesize a data distribution good enough to be mistaken by the real data distribution. However, optimizing GANs faces two challenges: mode collapsing and training instability. The former means that the generated data distribution concentrates on a small subset of outputs. The latter is due to the case that preserving a Nash equilibrium for both discriminator and generator is non-trivial. GAN variants~\cite{arjovsky2017wasserstein,gulrajani2017improved,mao2017least} propose improved theory to address these problems. Another line of work~\cite{Brock19,denton2015deep,karras2019style} investigates devising efficient architectures to generate high-resolution images.

\minisection{Transfer learning.}
This area aims to use the knowledge of the model (\ie, \textit{source}) trained on a large domain to accelerate the training and reduce the amount of training data required by a model (i.e., \textit{target}).  Related works study knowledge transfer on generative models~\cite{noguchi2019image,wang2018transferring, wang2020minegan, zhao2020leveraging,li2020few,wang2021minegan++,wang2020deepi2i,wang2021transferi2i} as well as discriminative models~\cite{donahue2014decaf}. Regarding generative models, TransferGAN~\cite{wang2018transferring} is one of the first works that explores transfer learning, using finetuning on pre-trained GANs and denoting good performance on small dataset.

\minisection{Hypernetworks.} 
Hypernetworks are implicit generators \cite{HaDL17,Ivan21} that aim to generate parameters for other models. 
Hypernetworks have  been applied to various tasks: architecture search \cite{ZhangRenUrtasun19}, few-shot learning \cite{Bertinetto16} and lifelong learning \cite{Oswald20}.  In this paper, we use Hypernetworks to generate the weights to modulate the learned weight of the pre-trained GAN. To our best knowledge, hypernetworks have not been used before to perform knowledge transfer. Furthermore, we use Hypernetworks to achieve the knowledge transfer from an unconditional GAN to a conditional GAN.
Our method can be seen as a straightforward implementation \cite{Oswald20} of a hypernetwork, producing the entire set of weights for a target neural network. However, we substitute the task embeddings $\{ \mathbf{e}^i \}_{i=0}^T$ for a semantically rich class space $\mathcal{V}$. Leveraging this space and the knowledge from the source domain of transfer learning, we are able to use more light-weight hypernetwork submodules, \ie, mainly consisting of simple affine transformations.

\minisection{Contrastive learning.}
In recent years, contrastive learning has been bridging the gap between supervised and unsupervised learning \cite{Chen20}. Data augmentation \cite{DoerschGE15,Zhirong18} has very often been used in representation learning to keep the mutual information of different augmentations while disregarding nuisances not useful for generalization. We can see it explicitly mixed into the GAN training dynamics \cite{Jure21,Shengyu20} when applied to the discriminator or the GAN objective as a form of data efficiency regularization.
Our application can be seen as a more simplified version of contraD \cite{Jeong21}, with a joint objective for real and fake samples, and SimCLR \cite{Chen20} is replaced with Barlow Twins \cite{Jure21} as the contrastive objective. To our knowledge, while some work has been carried out on augmentation \cite{chen2020hypernetwork} and semi-supervision \cite{brahma2021hypernetworks}, no other work has applied contrastive training to improve hypernetworks.

\section{Methodology}
We consider a source domain represented by the dataset $\mathcal{D}_s$ and a multi-class target domain $\mathcal{D}_t$.
Given a pre-trained model on the source domain $f_0(\cdot)$, we aim to use transfer learning to efficiently learn a hypernetwork $f_h(\cdot)$ that can generate weights for all classes of the target domain.

To shape an unconditional GAN into a conditional one, we introduce class specific parameters in \Cref{sec:weight-mod} that result in a certain modulation of the forward pass through the generator. This allows to drive the model toward the distributions of each target class.
Next, to prevent learning of separate modulation parameters for all the classes, in \Cref{sec:hypernetwork} we propose the hypernetworks to directly estimate the modulation parameters -- and importantly share the knowledge required to generate them among the classes. This is motivated by the fact that hypernetworks have been shown to efficiently transfer knowledge from one task to another one in the context of continual learning~\cite{Oswald20}.
However, since the introduced hypernetwork needs to be trained from scratch, the system suffers from hard optimization and long training. Thus, in \Cref{sec:self-init}, we present a new self-distillation method to learn  well-initialized weight for hypernetworks without the need of any data.  Finally, in \Cref{sec:barlow} we show that contrastive learning can be applied to further improve the efficiency of the knowledge transfer and improve the quality of the generation.

\subsection{Domain transfer} \label{sec:weight-mod}
Given a source generative model trained on $\mathcal{D}_s$, we aim to apply its knowledge to aid the learning of arbitrarily far domains.
Concretely, given a pre-trained (\ie, source domain) fully connected layer (or convolution, equivalently) $h^{\text{s}}(\boldsymbol{x}) = \boldsymbol{W} \boldsymbol{x} + \boldsymbol{b}$ with pre-trained weights $\boldsymbol{W} \in \mathbb{R}^{d_{\text{out}} \times d_{\text{in}}}$ and input $\boldsymbol{x} \in \mathbb{R}^{d_\text{in}}$. Inspired by~\cite{odena2017conditional,Perez18,Cong20}, we can modulate its statistics to form a different layer as
\begin{align}
    \label{eq:adafm}
    \hat{\boldsymbol{W}}_i &= \boldsymbol{\gamma}_i \odot \frac{\boldsymbol{W} - \boldsymbol{\mu}}{\boldsymbol{\sigma}} + \boldsymbol{\beta}_i, \\
    \hat{\boldsymbol{b}}_i &= \boldsymbol{b} + \boldsymbol{b}_i,
\end{align}
where $\boldsymbol{\gamma}_i, \boldsymbol{\beta}_i \in \mathbb{R}^{d_\text{out} \times d_\text{in}}$ are learned parameters, $i=1,...,N_c$ indicates the class, $N_c$ is the number of classes, and $\boldsymbol{\mu}, \boldsymbol{\sigma}$ are the mean and standard deviation of $\boldsymbol{W}_i$. The rationale behind this modulation is that it first removes the source style encoded in $\boldsymbol{\mu}, \boldsymbol{\sigma}$ and then apply the learned one from $\boldsymbol{\gamma}, \boldsymbol{\beta}$ to model the statistics of a generative process of the target distribution. This  normalization was originally proposed by~\cite{Cong20} and called \emph{Adaptive Filter Modulation} (AdaFM) in the context of continual learning of GANs.

\begin{figure}
    \centering
    \includegraphics[width=\linewidth]{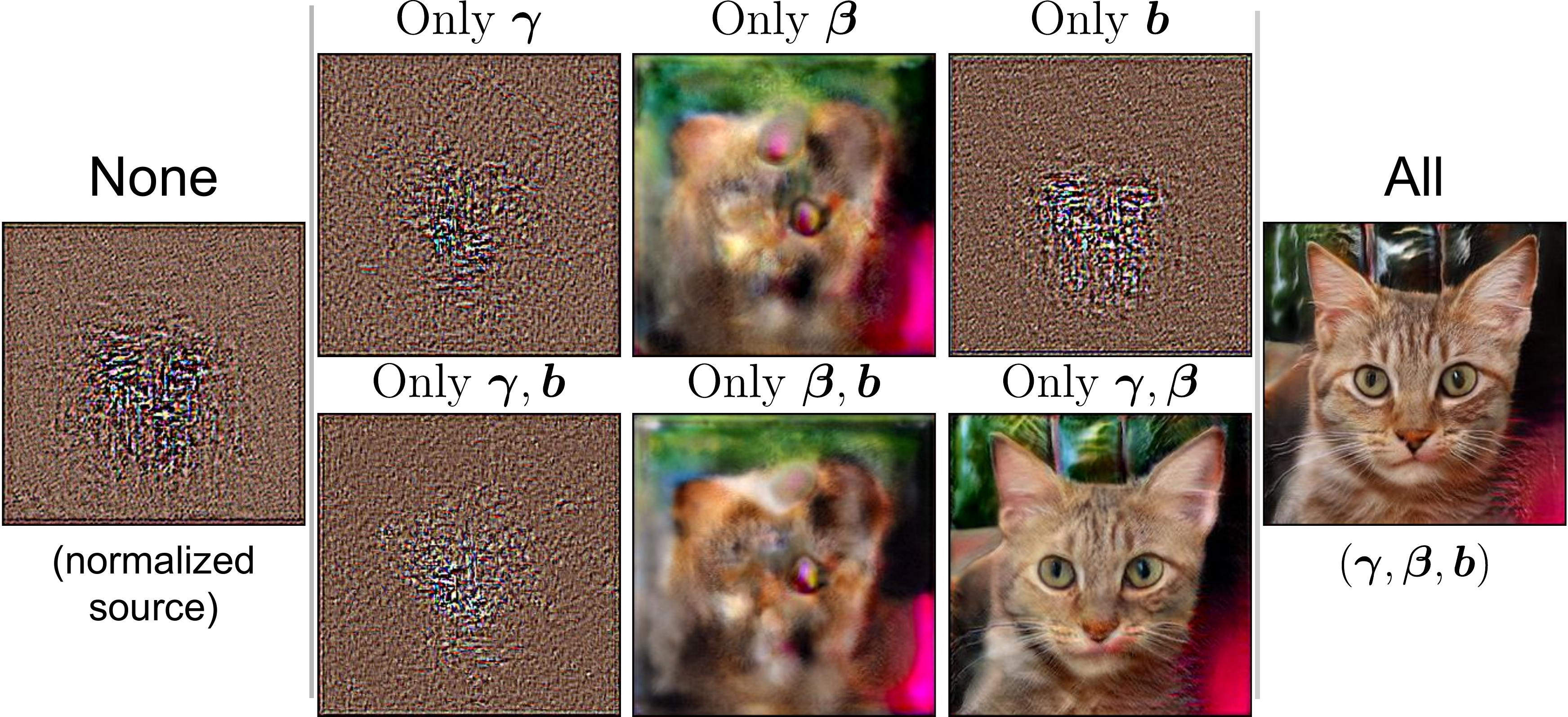}
    \caption{Effect of the modulation parameters on the domain transfer. Parameters $\boldsymbol{\gamma}$ generate high-frequency details, \ie, texture and structure. $\boldsymbol{\beta}$ takes care of low-frequency details, \ie, color. $\boldsymbol{b}$ is for localized details unattainable otherwise. 
    A detailed figure is shown in the Appendix \cref{fig:suppl:mod-params-big}}
    \label{fig:modulation-params}
\end{figure}

In another vein, we apply this modulation concurrently to tackle the problem of transfer learning to multiple domains. The network weights $\boldsymbol{W}$ and $\boldsymbol{b}$ are shared among all the transferred classes, while the modulation parameters $\boldsymbol{\gamma}, \boldsymbol{\beta}, \boldsymbol{b}$ are the only ones changing. In \Cref{fig:modulation-params} we can see the effect of each parameter in the knowledge transference. In~\cite{Cong20} they show that this modulation allows to model large domain shifts.
Conditioning $\boldsymbol{\gamma}, \boldsymbol{\beta}$ and $\boldsymbol{b}$ we will be able to harness the modulated generation to produce conditional networks from an unconditional base.

\subsection{Hyper-modulation} \label{sec:hypernetwork}

The method proposed in the previous section (Eq.~\ref{eq:adafm}) is optimized for each class in the target domain separately, and no parameters of the modulation are shared among the classes.
As a result, we do not exploit similarities among classes in the target domains.
To solve this, we propose the usage of hypernetworks~\cite{HaDL17}, allowing us to share information and reduce memory usage by accumulating knowledge in the newly introduced modules.

Neural networks $f(\mathbf{x}, \Theta)$ are a family of functions that, given an input $\mathbf{x}$ and an output $\mathbf{y}$ coming from a dataset 
$\mathcal{D} = \{ (\mathbf{x}, \mathbf{y}) \}$,
typically learn a set of parameters $\Theta$ to find a function that maximizes the log likelihood of the data. Hypernetworks \cite{HaDL17,Oswald20} aim to learn the parameters $\Theta_h$ of a metamodel, which then will generate the target parameters $\Theta_\text{trg}$ of the target model $f_\text{trg}$.

\begin{figure}
    \centering
    \includegraphics[width=\linewidth]{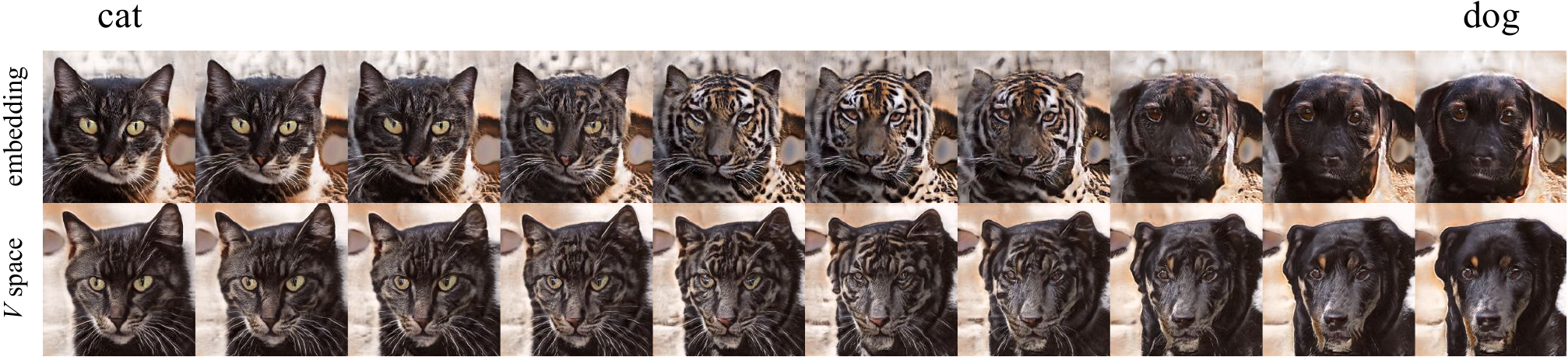}
    \caption{Conditioning interpolation on the modulation. Introducing a class projector on $\mathcal{V}$ results in smoother interpolation, although some features of other classes keep leaking while traversing. Additional interpolations can be found in the Appendix.}
    \label{fig:latent-interpolation}
\end{figure}

In this work, we apply a hypernetwork $g$ to predict the modulation parameters conditionally, which eventually enables us to produce a generative model for each target. The input of the hypernetwork is a vector coming from a class embedding network  $C(i;\Psi) = \mathbf{v}\in\mathcal{V}$ where ($i=1,...,N_c$) is the class label, $\mathcal{V}$ is the class embedding space, and $\Psi$ are network parameters. \Cref{fig:latent-interpolation} shows qualitative improvement over learnable embeddings, and the Appendix includes metrics and more extensive visualizations. By varying the number of parameters $\Psi$, we are able to vary the class knowledge capacity of the system. The hypernetwork $g$ takes the embedding vector $v$ and maps it to the modulation parameters according to:
\begin{align}
    \boldsymbol{\gamma}_\mathbf{v}, \boldsymbol{\beta}_\mathbf{v} &= g(\mathbf{v};\Phi_a), \qquad \boldsymbol{b}_\mathbf{v} = g_b(\mathbf{v};\Phi_b)
    \label{eq:parametrized-adafm}
\end{align}
where $g$ are affine projections of a point in the space $\mathcal{V}$, with network parameters $\Phi_a$ and $\Phi_b$. We use $\mathbf{\Phi}$ to denote the combination of all the parameters used by the hypernetwork, consisting of $\Phi_a$ and $\Phi_b$ for all the layers in the network. Each modulated layer has a $g$ projector, but layer-wise, these are shared among target classes.

\begin{figure}
    \centering
    \includegraphics[width=0.9\linewidth]{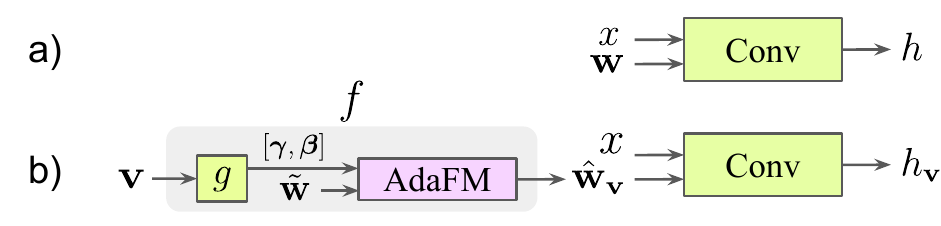}
    \caption{a) Activations $h$ from a generator convolution in the source domain. b) Domain-specific activations from a hypermodulator $f$, conditioned on a point $\mathbf{v}$ in the class space $\mathcal{V}$.
    \label{fig:conv-block}}
\end{figure}

The modulation that produces target-specific activations $h_\mathbf{v}(\boldsymbol{x}) = \hat{\boldsymbol{W}}_\mathbf{v} \boldsymbol{x} + \hat{\boldsymbol{b}}_\mathbf{v}$ is of the form
\begin{align}
    \label{eq:adafm2}
    \hat{\boldsymbol{W}}_\mathbf{v} &= \boldsymbol{\gamma}_\mathbf{v} \odot \frac{\boldsymbol{W} - \boldsymbol{\mu}}{\boldsymbol{\sigma}} + \boldsymbol{\beta}_\mathbf{v} , \\
    \hat{\boldsymbol{b}}_\mathbf{v} &= \boldsymbol{b} + \boldsymbol{b}_\mathbf{v} ,
\end{align}
where $\boldsymbol{W}$ and $\boldsymbol{b}$ are the frozen source weights.
Ultimately, a hypermodulator $f$
will be given a class embedding $\mathbf{v}$ and a normalized source weight $\tilde{\mathbf{w}}$ to produce the desired target weights as $f_{\tilde{\mathbf{w}}}(\mathbf{v}) = \boldsymbol{\gamma}_\mathbf{v} \odot \tilde{\mathbf{w}} + \boldsymbol{\beta}_\mathbf{v} = \hat{\mathbf{w}}_\mathbf{v}$, following \cref{eq:parametrized-adafm,eq:adafm2} and pictured in \cref{fig:conv-block}.

Traditionally, reusability can be introduced in hypernetworks to reduce the number of trainable parameters. This is achieved by reapplying the metamodel for different partitions of the target model parameters, also called chunking \cite{Oswald20}.
We do not use chunking since each generator can be reduced to a minimum of a learned affine transformation thanks to transfer learning and the enhanced domain space $\mathcal{V}$, constituting a rather shallow but performing hypernetwork.

\subsection{Self-alignment} \label{sec:self-init}

The introduction of the new modules causes the augmented source model to initially lose its learned synthesis performance (see also \cref{fig:self-init-start}), 
mainly because the parameters $\Psi,\Phi$ have not been learned yet, as well as due to the removal of domain-specific statistics prior to the introduction of new ones, as seen in \cref{eq:adafm2}. This procedure is not necessarily bad, since new classes will only learn to produce their respective target statistics and not to compensate for the source ones. However, general training times will be affected since the network has to re-learn multi-scale feature statistics that produce real-world pixel distributions.

\begin{figure}
    \centering
    \includegraphics[width=0.65\linewidth]{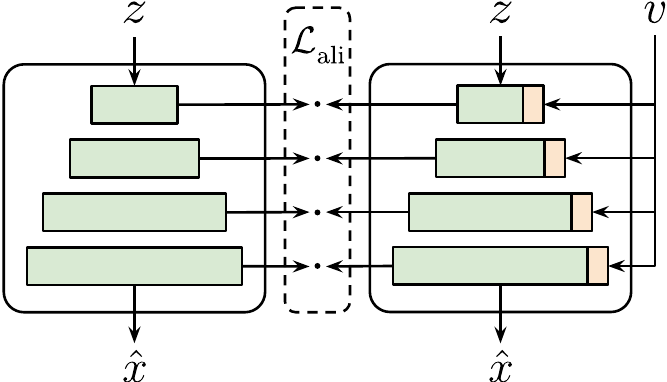}
    \caption{Self-alignment of the pre-trained generator (left) and the one with hypernetwork (right). Both networks are initialized with the same pre-trained weights (green) that are frozen. The new hypernetwork weights (yellow) are learned during the self-alignment. This operation does not require any data, since it can be performed by simply sampling a latent vector $z$.}
    \label{fig:alignment}
\end{figure}

Therefore, we propose to self-align the parameters $\Psi,\Phi$. The alignment is performed between the pre-trained generator network without hypernetwork and the one with hypernetwork (see ~\cref{fig:alignment}). The aim is to not simply recover the original weight statistics, but also to initialize a sensible latent space for the embedding vectors $v$ that could be further augmented by new classes.

We will perform this initialization as a first step before the final finetuning on the target data takes place. The hierarchical features extracted from the pre-trained model are given by $F_\text{PT}(\mathbf{z}) = \{ G_\text{PT}(\mathbf{z})_l \}$ and the ones with hypernetwork by $F_\text{hyp}(\mathbf{z}) = \{ G'_\text{PT}(\mathbf{z}, g(C(c^0;\Psi);\Phi))_l \}$ where $G(\cdot)_l$ is the $l$-th convolution block output. During the self-initialization, we set the class input to the class-embedding network $C$ as $c^0=1$. The loss for this stage is:
\begin{equation}
    \mathcal{L}_\text{ali} = \sum_l \lVert F_\text{PT}(\mathbf{z}) - F_\text{hyp}(\mathbf{z}) \rVert_1 .
\end{equation}
Note that this operation does not require any real data, since we can align the two networks by simply sampling random vectors $\mathbf{z}$. After self-initialization, the network with the hypernetwork generates high-quality images (compare~\cref{fig:self-init-trained} and \cref{fig:self-init-gt}).

In conclusion, the self-alignment initializes the hypernetwork parameters $\Psi,\Phi$. When we now finetune the network on the multi-class target domain, we do not have to learn these parameters from scratch. In the experimental section, we verify that this significantly reduces the training time and improves the quality of the generated results.

\subsection{Contrastive learning} \label{sec:barlow}
We further extend this work to achieve better sample efficiency by applying contrastive learning on the discriminator used during adversarial training.
Recent works on self-supervised learning have shown that by mapping different views (generated by taking different data augmentations of the same image) to the same point in latent space, strong semantically-rich feature representations can be learned that rival their supervised counterparts. Here, the idea is to exploit this fact to improve the quality of the discriminator used in adversarial training. The underlying insight is that if the discriminator can extract higher quality features, it can also better distinguish fake from real images, and as a consequence better challenge the generator, leading to higher quality images. 

\begin{figure}
    \centering
    \includegraphics[width=\linewidth]{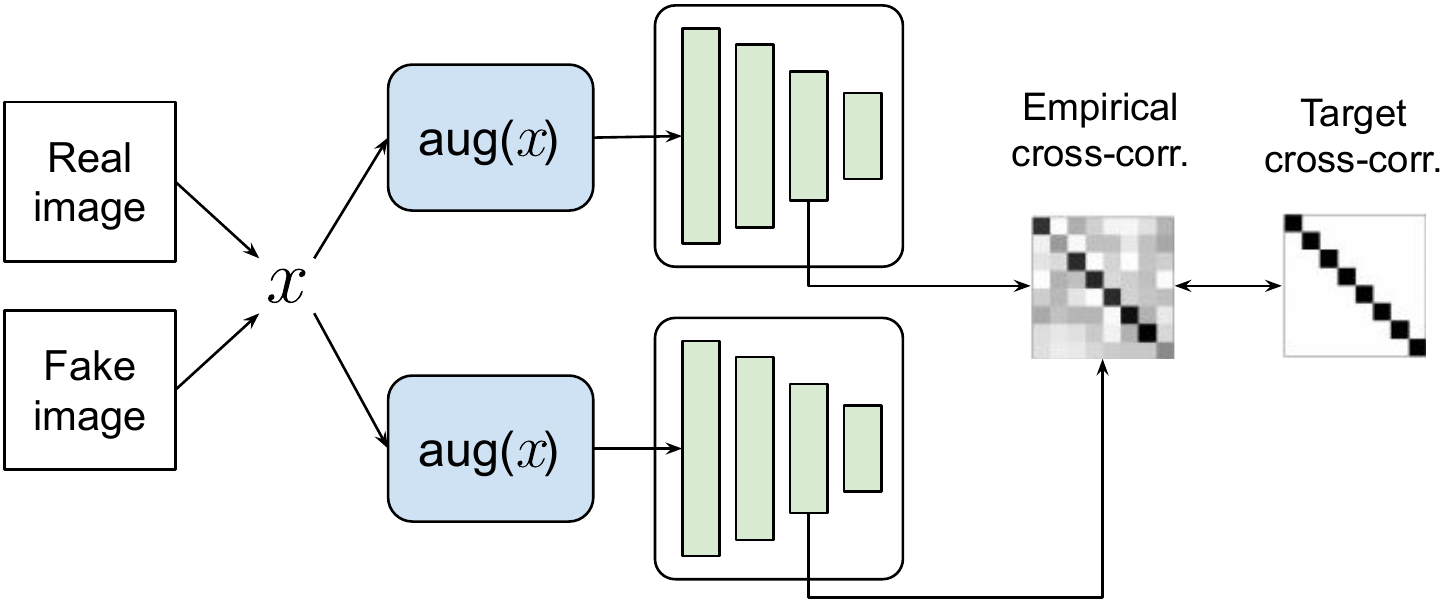}
    \caption{General scheme for contrastive learning on the discriminator. Cross-correlation (\textit{cross-corr.}) is computed between the extracted features of two views of a real or fake image. Gradients don't flow back to the generator.}
    \label{fig:barlow}
\end{figure}

Concretely, we make use of Barlow Twins \cite{Jure21} for its simplicity and performance and apply it implicitly on the discriminator as in \cref{fig:barlow}. We reuse all transformations for real and fake samples, but we employ no projector network because it resulted in worse quality. The loss function is also left unchanged:
\begin{equation}
    \mathcal{L}_\text{contr} = \sum_i (1 - \mathcal{C}_{ii})^2 + \lambda \sum_i \sum_{j \ne i} {\mathcal{C}_{ij}}^2
\end{equation}
with the scaling factor $\lambda$ and the cross-correlation matrix $\mathcal{C}$ computed between the intermediate representations  before the final layer.

We employ GAN~\cite{goodfellow2014generative}  to optimize this problem:
\begin{equation}
\begin{aligned}
\mathcal{L}_{gan} &= \mathbb{E}_{\mathbf{x}\sim \mathcal{X}, \mathbf{c} \sim p(\mathbf{c})}\left[ \log D \left ( \mathbf{x}, \mathbf{c} \right ) \right]  \\
&+   \mathbb{E}_{\mathbf{z} \sim p(\mathbf{z}), \mathbf{c} \sim p(\mathbf{c})}\left[ \log (1 - D \left ( G \left (\mathbf{z}, \mathbf{c} \right ), \mathbf{c}  \right ) \right],
\end{aligned}
\end{equation} 
where $p \left ( \mathbf{z} \right )$ follows the normal distribution, and $\mathbf{p} \left ( \mathbf{c} \right )$ is the domain label distribution.

The final training objective is
\begin{equation}
    \mathcal{L}_\text{GAN} =\mathcal{L}_{gan} + \lambda_\text{contr} \mathcal{L}_\text{contr}
\end{equation}
where $\lambda_\text{contr}$ is a balancing hyperparameter set to $\lambda_\text{contr}=1e-3$ in all our experiments. In the experimental section, we verify that contrastive learning can significantly improve the quality of the generated images. 

\section{Experiments}

\subsection{Settings}
\minisection{Training details.} Our method is applied to a pre-trained StyleGAN \cite{karras2019style}. Concretely, both the generator and discriminator are direct copies of the architecture, except for the top layer of the discriminator, for which the last fully connected layer has been replaced by a convolutional layer with $3 \times 3$ filter size, stride of 1 and output channel dimensionality of $N_c$ number (classes in target domain).
The hypernetwork class network $C(\cdot)$ consists of an embedding layer for all domains, followed by four fully connected layers. The dimensionality of the whole branch is 64. The hypernetwork modulators are implemented by a single fully connected layer that maps the class branch output to a dimensionality of 512.
Hyperparameters from the original model are kept, including Adam~\cite{kingma2014adam} and R1 regularization \cite{DBLP:journals/corr/abs-1801-04406}, while the model is trained at $256 \times 256$ resolution.

\minisection{Evaluation metrics.}
We report results on two types of metrics: single-valued and double-valued metrics. The former contains Fr\'echet Inception Distance (FID)~\cite{heusel2017gans} and Kernel Inception Distance (KID)~\cite{binkowski2018demystifying}. The latter consists of Precision and Recall (PR)~\cite{kynkaanniemi2019improved} and Density and Coverage (DC)~\cite{naeem2020reliable}. Both PR and DC evaluate the quality and the diversity. We use all training samples available to compute the metrics as suggested in \cite{heusel2017gans,karras2020training}, since most datasets do not have as much as 10,000 class samples per class to have a good metric estimation. KID and DC are multiplied by 100 for easier visualization, and PR is given as percentage. FID is calculated per class and the average is taken (mFID).

\minisection{Datasets.} Our experiments are  conducted on Animal Faces dataset (AFHQ)~\cite{choi2020stargan}, FFHQ~\cite{karras2019style}, CelebA-HQ \cite{karras2017progressive},  Flowers102~\cite{nilsback2008automated} and Places 365~\cite{zhou2017places}.  AFHQ contains 3 classes, each one has about 5000 images.  In CelebA-HQ, we use gender as a class, with 10k male and 18k female images in the training set.  Flowers102 consists of 102 categories, but since the number of samples per class is small, we ignore the labels to form an unconditional dataset. In Places 365~\cite{zhou2017places} dataset, we select only 10 categories as target: \textit{amphitheater}, \textit{aqueduct}, \textit{castle}, \textit{dam}, \textit{field road}, \textit{fire station}, \textit{pagoda}, \textit{underwater - ocean deep}, \textit{volcano} and \textit{waterfall}. In this paper, all images are resized to $256 \times 256$.

\minisection{Baselines.} 
Since no previous work has explored transfer learning from unconditional to conditional GANs, there exist only few works to properly compare against. We use the following baselines:  
\textit{GAN Memory}~\cite{Cong20} (unconditional to unconditional) proposed a weight modulation method to address catastrophic forgetting of GAN for lifelong learning, \textit{cGANTransfer}~\cite{Shahbazi_2021_CVPR} (conditional to conditional) introduced a conditional batch normalization method to perform knowledge transfer, which aims to learn the class-specific information of the new classes from that of the old classes. 
We explore a variant of our method, named as \textit{Hyper-Mod-FT}, for which all parameters are updated.

\subsection{Ablation study}
\begin{table}[t]
  \centering
  \resizebox{1\columnwidth}{!}{
  \begin{tabular}{@{}lcccccc@{}}
    \toprule
    Configuration & mFID $\downarrow$ & mKID $\downarrow$ & P $\uparrow$ & R $\uparrow$ & D $\uparrow$ & C $\uparrow$ \\
    \midrule
    \quad \, No hypernet. & 61.84 & 3.75 & 8.89 & 22.77 & 2.37 & 1.33 \\
    \midrule
    {\small A} \, Hyper-Mod & 50.67 & 3.00 & 12.46 & 31.60 & 3.58 & 3.03 \\
    {\small B} \, + $\mathcal{V}$ space & 45.28 & 2.28 & 12.12 & 40.18 & 3.67 & 3.31 \\
    {\small C} \, + Contrastive D & 26.74 & 0.92 & 28.02 & 55.19 & 10.13 & 11.95  \\
    \bottomrule
  \end{tabular}
  }
  \caption{Ablation on hypernetwork and the contrastive learning on AFHQ. }
  \label{tab:ablation}
\end{table}

\minisection{Hypernetwork.}
Comparing a modulation like \textit{GAN Memory} (No hypernet.) to the proposed hypernetwork (config.\ {\small A}) in \Cref{tab:ablation},
we can appreciate better synthesis quality and especially a diversity increase for the latter, more than doubling for both Recall and Coverage. We attribute that to the knowledge sharing and complementary supervision in the joint training, since each input is affecting and shaping the whole hypernetwork as opposed to learning separate embedding points for modulation.

\begin{figure*}
\begin{subfigure}{0\textwidth}
\refstepcounter{subfigure}\label{fig:self-init-start}
\end{subfigure}%
\begin{subfigure}{0\textwidth}
\refstepcounter{subfigure}\label{fig:self-init-trained}
\end{subfigure}
\begin{subfigure}{0\textwidth}
\refstepcounter{subfigure}\label{fig:self-init-gt}
\end{subfigure}
\begin{subfigure}{0\textwidth}
\refstepcounter{subfigure}\label{fig:short-b}
\end{subfigure}

  \centering
    \includegraphics[width=\textwidth]{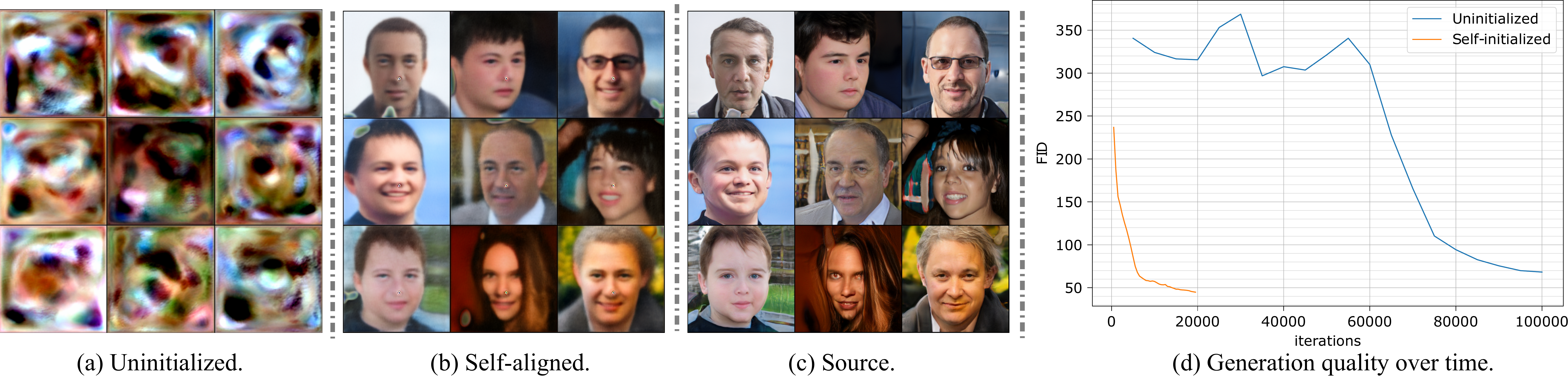}
  \caption{Self-initialization details. Generator outputs of a hypermodulator (a, b). Source pre-trained generator (c) for comparison. Training efficiency of self-alignment (d).}
  \label{fig:self-align-2}
\end{figure*}

\minisection{Self-initialization effect.}
Training with an uninitialized hypernetwork (\cref{fig:self-init-start}) is compared to a self-aligned one (\cref{fig:self-init-trained}) towards a source model (\cref{fig:self-init-gt}). \Cref{fig:short-b} shows huge improvements in training time as well as a significant improvement in quality. We argue that learning proper hierarchical modulation correlation plays a crucial role for consecutive direct application of training information to each target, compared to learning both concurrently from scratch.

\minisection{Target space.}
A commonly desired characteristic of latent spaces is the linearity of its factors of variation (\eg pose, color, etc.). Our goal with the class network \textit{C} introduced in \Cref{sec:hypernetwork} is to unwarp subspaces that the learned class embedding could have had difficulties dealing with for several reasons, \ie scarcity of specific training samples or complexity of the modelling.
To quantify the beneficial effect of the introduced module, we employ a disentanglement metric called Perceptual path length \cite{karras2019style}, consisting on measuring how drastic perceptual changes in the image occur while performing interpolation. Intuitively, a linear latent space presents smoother transitions than a warped one.
Results shown in \Cref{tab:suppl:ppl} confirm us the advantage of introducing this network against class embeddings. Magnitudes are naturally bigger than style measurements since changes in class are non-trivial perceptual alterations compared to, \eg, color changes.
\begin{table}
  \centering

  \begin{tabular}{@{}lcc@{}}
    \toprule
    \quad Method & full & end \\
    \midrule
    {\small A} \, Hyper-Mod & 61.94 & 61.78 \\
    {\small B} \, + $\mathcal{V}$ space & \textbf{61.23} & \textbf{59.85} \\
    \bottomrule
  \end{tabular}
  \caption{Perceptual path length among classes, for both full paths and endpoints. All scores are in the magnitude of $10^6$.}
  \label{tab:suppl:ppl}
\end{table}
Generation metrics also denote improvement in quality and diversity in \Cref{tab:ablation} (config.\ {\small B}).
Finally, we show in \Cref{fig:suppl:interpolation_class_noise} that class regarding style has appropriate independence, \ie, changes in class only affect shape, but fur color, background, etc.\ are left unchanged. Style regarding class cannot be dependent since the style mechanism is frozen at the beginning of the transfer.

\minisection{Contrastive learning.}
We found self-supervision beneficial to transfer learning. We tried some contrastive losses (see \Cref{tab:others}) and choose the best one. This provides improvements even with a small batch size of 10 samples, for which we compute an FID of 37.15 and KID of 1.66, already improving config.\ {\small B} in \Cref{tab:ablation}. Results reported in config.\ {\small C} are computed for a batch size of just 60 due to computational constraints.
These are expected to further improve with larger batch sizes, as in \cite{Jure21}.
Unfortunately, contrastive learning in the generator did not result in improved quality. 

\minisection{Domain information injection.} 
Is weight modulation the best method to incorporate target information during the transfer learning? We could think about style transfer techniques such as Adaptive Instance Normalization (AdaIN) \cite{karras2019style}, which modulates at the activation level.
In \Cref{sec:suppl:domain_mod_method} we provide specifications on the implementation of this method in place of modulation.
This modification can be compared to config.\ {\small B} and yields an FID and KID of 110.55 and 9.26 respectively (compared to our proposed architecture with 45.28 and 2.28). Thus, we conclude that weight modulation is favorable over other style transfer methods.

\subsection{Result}

\begin{table}[t]

    \setlength{\tabcolsep}{1mm}
    \resizebox{\columnwidth}{!}{%
    \centering
    \footnotesize
    \setlength{\tabcolsep}{1pt}
    \begin{tabular}{|c|c|c|c|c|}
    \hline
    \multirow{2}{*}{\diagbox{Method}{Dataset}} &\multicolumn{2}{c|}{Close domain }&\multicolumn{2}{c|}{Far domain}\cr\cline{2-5}
      & FFHQ$\rightarrow$AFHQ & AFHQ$\rightarrow$CelebA & FFHQ$\rightarrow$Flower102 & FFHQ$\rightarrow$Places365 \cr\cline{2-5}
      \hline
  Hyper-Mod-S & 498.41 & 498.41 & 498.41 & 498.41 \cr\cline{2-5}
     \hline
    GAN Memory & 61.84 & 49.30 & 144.93 & 229.49 \cr\cline{2-5}
    \hline
    cGANTransfer & 112.64 & 105.95 & - & - \cr\cline{2-5}
    \hline
    Hyper-Mod-FT & 30.11 & 24.54 & 40.07 & 98.24 \cr\cline{2-5}
 \hline
    Hyper-Mod & 45.28 &  45.54 & 127.78 & 132.42  \cr\cline{2-5}
    \hline
    \end{tabular}  
    }
  \caption{Comparison with baselines on mean FID. A$\rightarrow$B: From source A to target B. S: From scratch. FT: finetune source weights.}
  \label{tab:results:quantitative}
\end{table}
\begin{table}[t]
  \centering
 \resizebox{1\columnwidth}{!}{
    \setlength{\tabcolsep}{10pt}

  \begin{tabular}{@{}lcccccc@{}}
    \toprule
    Configuration & mFID $\downarrow$ & mKID $\downarrow$ & P $\uparrow$ & R $\uparrow$ & D $\uparrow$ & C $\uparrow$ \\
    \midrule
    GAN Memory \cite{Cong20} & 61.84 & 3.75 & 8.89 & 22.77 & 2.37 & 1.33 \\
    cGANTransfer \cite{Shahbazi_2021_CVPR} & 112.64 & 9.90 & 2.93 & 18.95 & 0.73 & 2.10 \\
    Hyper-Mod & 45.28 & 2.28 & 12.12 & 40.18 & 3.67 & 3.31 \\
    Hyper-Mod-FT & 30.11 & 1.09 & 16.99 & 62.68 & 5.76 & 6.49 \\
    \midrule
    Hyper-Mod + DCL \cite{yu2021dual} (bs 60) & 42.28 & 2.00 & 19.82 & 42.05 & 7.31 & 5.67 \\
    Hyper-Mod + BT \cite{Jure21} (bs 60) & 26.74 & 0.92 & 28.02 & 55.19 & 10.13 & 11.95 \\
    \bottomrule
  \end{tabular}
  }
  \caption{Comparison with baselines on several metrics on AFHQ. P: Precision, R: Recall, D: Density and C: Coverage.}
  \label{tab:others}
\end{table}

\minisection{Quantitative results}. To evaluate the performance of the proposed method, we test our method on both \textit{close domain transfer} and \textit{far domain transfer}. The former means both source and target domain have small domain shift, and the latter is they have a large domain gap. These two settings are used to validate the effectiveness of the proposed method on different target domains. 

\begin{figure*}[t]
    \centering
    \includegraphics[width=0.95\linewidth]{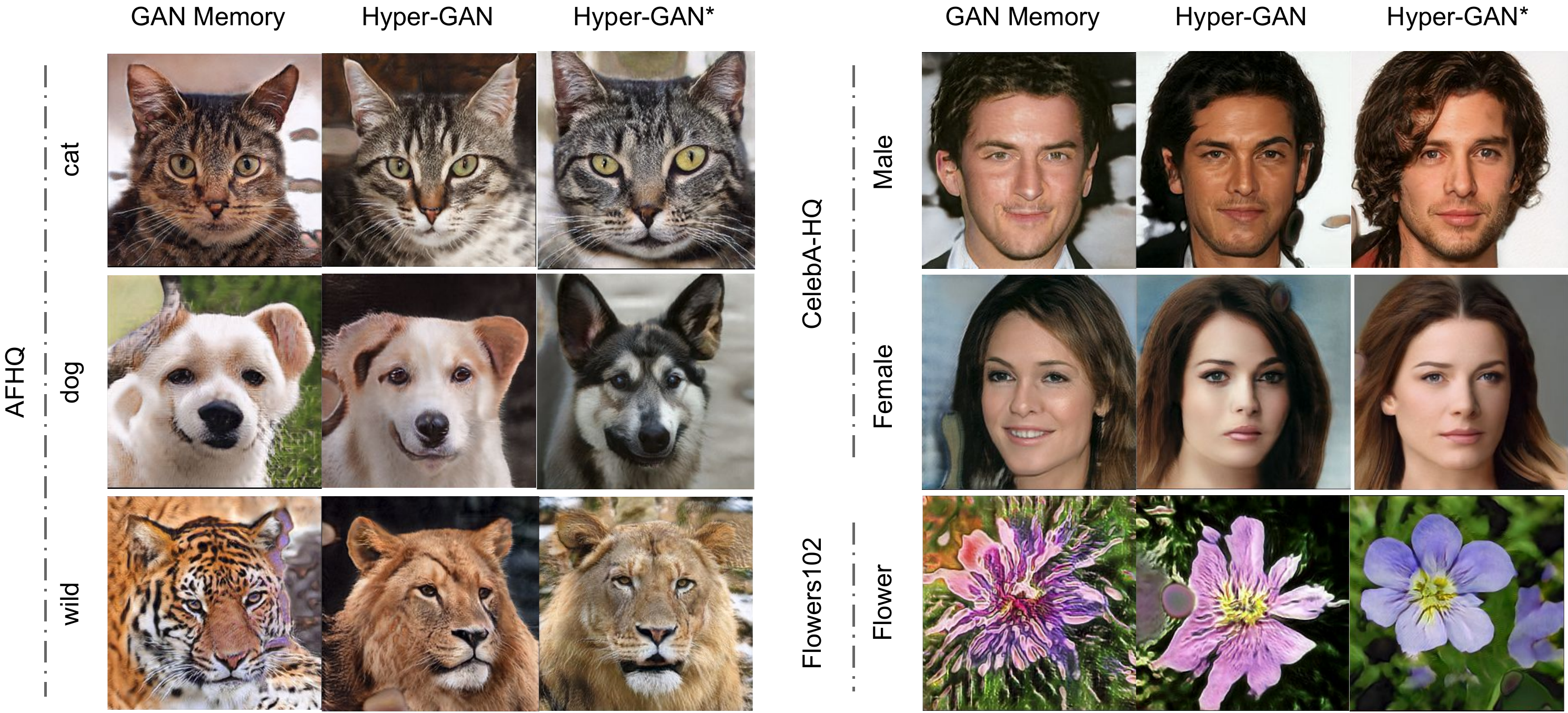}
    \caption{ Qualitative comparison on  AFHQ, CelebA-HQ and Flowers102 datasets.  More examples are shown in the Suppl. Mat. Section.}
    \label{fig:qualitative:results}
\end{figure*}

\begin{figure}[t]
    \centering
    \includegraphics[width=\linewidth]{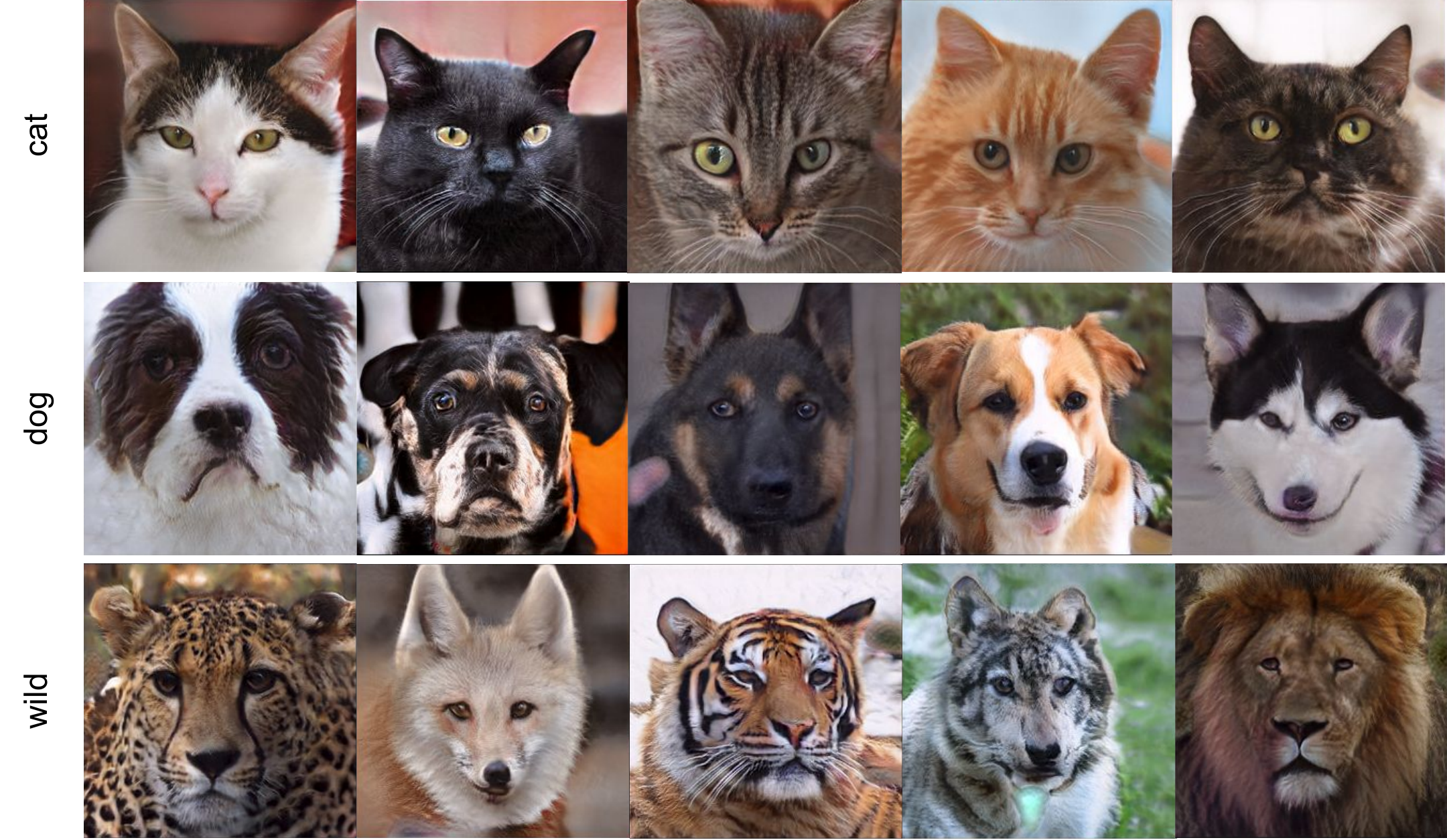}
    \caption{ Qualitative results of the proposed method on AFHQ. Each row is corresponding to one target class. For each target, we show five different breeds appearing in the generation.}
    \label{fig:qualitative:diversity}
\end{figure}

\minisection{\textit{Close domain transfer}}. Here, we use both the AFHQ animal dataset and CelebA human face as our target domains.  For the former, the pretrained StyleGAN is optimized on FFHQ human face. We use the pretrained StyleGAN optimized on AFHQ animal face when the target domain is CelebA human face.  As reported in \Cref{tab:results:quantitative} (\textit{close domain} column), training the network from scratch obtains catastrophic results (e.g., 498.41 FID). Using the transfer learning method (like GAN Memory) largely improves the performance (e.g., 61.84 FID for GAN Memory).    
The proposed method achieves better performance (denoted as \textit{Hyper-Mod} in the table), we generate more realistic and correct class-specific images among the compared methods. %
In addition, we also conduct an experiment with updating all parameters (denoted as \textit{Hyper-Mod-FT}). \textit{Hyper-Mod-FT} further improves the performance.

We also evaluate our method and the baselines  on several other metrics. As reported in Table~\ref{tab:others}, we achieve the best score on all metrics, which indicates that we not only generate high-quality images (corresponding to P. and D.), but also diverse images (corresponding to R. and C.).

\minisection{\textit{Far domain transfer.}}  We also consider the challenging setting by using a target dataset which has a large domain gap with the source domain. Here we consider two target domains: Flower102 and Places365. For the two target datasets, we use the same source pre-trained StyleGAN, which is optimized on FFHQ. As reported in \Cref{tab:results:quantitative} (\textit{far domain} column), in the far domain setting our method still obtains a large advantage when compared to the baselines (e.g., 127.78 FID (ours) vs 144.93 FID (GAN Memory) on Flower102). What is more interesting is that we are able to 
greatly improve the performance when further updating all parameters. Finally, like for the close domain transfer, the proposed techniques (\ie, hypernetwork, self-alignment and contrastive learning)  are effective when performing knowledge transfer from unconditional GAN to conditional GAN on far domain transfer. 

\minisection{Qualitative results.}
Regarding \textit{close domain transfer}, \Cref{fig:qualitative:results} shows the comparison to baselines on AFHQ, CelebA and Flowers102 datasets.  Although GAN Memory is able to conduct multi-class generation, it fails to generate highly realistic images (first column of \Cref{fig:qualitative:results} on AFHQ).  Taking AFHQ as an example, given the target class label, the proposed method is able to provide high-quality images (e.g., the second column  of \Cref{fig:qualitative:results}). when updating all parameters (Hyper-Mod-FT), we further improve the qualitative result (e.g., the third column  of \Cref{fig:qualitative:results}).
Moreover, we demonstrate that our method has both scalability and diversity in a single model. Each row of \Cref{fig:qualitative:diversity} shows the different results when changing the target class label. Our method manages to cover different breeds in the same class, while keeping source style controls (\ie, colors, pose, background, etc.) unaltered (\cref{fig:latent-interpolation}).

For \textit{far domain transfer}, qualitative results on \textit{Places365} dataset to complement quantitative ones can be seen in Appendix \Cref{fig:suppl:places}, together with additional unfiltered generations and interpolations.

\section{Conclusions}
We investigated the knowledge transfer from GAN to cGAN. To tackle it, we proposed hyper-modulation to produce weight modulation parameters on-the-fly for a source model. Training the hypernetwork from scratch complicates training, thus we proposed a self-initialization method that does not require any data to learn well-initialized weights. To enhance the capacity of the discriminator, we introduced self-supervision for it. Our qualitative and quantitative results showed the proposed method outperforms existing state-of-the-art results on transfer learning.

\minisection{Limitations}
One further line of work is memory, to not keep the whole pre-trained network in memory. Second, the state of the art in unconditional generation uses a similar modulation \cite{karras2020analyzing} to incorporate the style. We are positive that combining and leveraging both methods is possible.

\section*{Acknowledgements}
We acknowledge the support from Huawei Kirin Solution, Spain Government funded project PID2019-104174GB-I00/AEI/10.13039/501100011033 and the EU Project CybSpeed MSCA-RISE-2017-777720.

{\small
\bibliographystyle{ieee_fullname}
\bibliography{shortstrings,refs}

\begin{thebibliography}{10}\itemsep=-1pt

\bibitem{arjovsky2017wasserstein}
Martin Arjovsky, Soumith Chintala, and L{\'e}on Bottou.
\newblock Wasserstein gan.
\newblock In {\em ICLR}, 2017.

\bibitem{Bertinetto16}
Luca Bertinetto, Jo{\~{a}}o~F. Henriques, Jack Valmadre, Philip H.~S. Torr, and
  Andrea Vedaldi.
\newblock Learning feed-forward one-shot learners.
\newblock In {\em NeurIPS}, 2016.

\bibitem{binkowski2018demystifying}
M Bi{\'n}kowski, DJ Sutherland, M Arbel, and A Gretton.
\newblock Demystifying mmd gans.
\newblock In {\em ICLR}, 2018.

\bibitem{brahma2021hypernetworks}
Dhanajit Brahma, Vinay~Kumar Verma, and Piyush Rai.
\newblock Hypernetworks for continual semi-supervised learning.
\newblock {\em arXiv preprint arXiv:2110.01856}, 2021.

\bibitem{Brock19}
Andrew Brock, Jeff Donahue, and Karen Simonyan.
\newblock Large scale {GAN} training for high fidelity natural image synthesis.
\newblock In {\em ICLR}, 2019.

\bibitem{chen2020hypernetwork}
Chih-Yang Chen and Che-Han Chang.
\newblock Hypernetwork-based augmentation.
\newblock {\em arXiv preprint arXiv:2006.06320}, 2020.

\bibitem{Chen20}
Ting Chen, Simon Kornblith, Mohammad Norouzi, and Geoffrey~E. Hinton.
\newblock A simple framework for contrastive learning of visual
  representations.
\newblock In {\em ICML}, 2020.

\bibitem{choi2020stargan}
Yunjey Choi, Youngjung Uh, Jaejun Yoo, and Jung-Woo Ha.
\newblock Stargan v2: Diverse image synthesis for multiple domains.
\newblock In {\em CVPR}, 2020.

\bibitem{Cong20}
Yulai Cong, Miaoyun Zhao, Jianqiao Li, Sijia Wang, and Lawrence Carin.
\newblock {GAN} memory with no forgetting.
\newblock In {\em NeurIPS}, 2020.

\bibitem{denton2015deep}
Emily~L Denton, Soumith Chintala, Rob Fergus, et~al.
\newblock Deep generative image models using a laplacian pyramid of adversarial
  networks.
\newblock In {\em NeurIPS}, pages 1486--1494, 2015.

\bibitem{DoerschGE15}
Carl Doersch, Abhinav Gupta, and Alexei~A. Efros.
\newblock Unsupervised visual representation learning by context prediction.
\newblock In {\em ICCV}, pages 1422--1430, 2015.

\bibitem{donahue2014decaf}
Jeff Donahue, Yangqing Jia, Oriol Vinyals, Judy Hoffman, Ning Zhang, Eric
  Tzeng, and Trevor Darrell.
\newblock Decaf: A deep convolutional activation feature for generic visual
  recognition.
\newblock In {\em ICML}, pages 647--655, 2014.

\bibitem{goodfellow2014generative}
Ian Goodfellow, Jean Pouget-Abadie, Mehdi Mirza, Bing Xu, David Warde-Farley,
  Sherjil Ozair, Aaron Courville, and Yoshua Bengio.
\newblock Generative adversarial nets.
\newblock In {\em NeurIPS}, pages 2672--2680, 2014.

\bibitem{gulrajani2017improved}
Ishaan Gulrajani, Faruk Ahmed, Martin Arjovsky, Vincent Dumoulin, and Aaron~C
  Courville.
\newblock Improved training of wasserstein gans.
\newblock In {\em NeurIPS}, pages 5767--5777, 2017.

\bibitem{JMLR:v13:gutmann12a}
Michael~U. Gutmann and Aapo Hyv{{\"a}}rinen.
\newblock Noise-contrastive estimation of unnormalized statistical models, with
  applications to natural image statistics.
\newblock {\em Journal of Machine Learning Research}, 13(11):307--361, 2012.

\bibitem{HaDL17}
David Ha, Andrew~M. Dai, and Quoc~V. Le.
\newblock Hypernetworks.
\newblock In {\em ICLR}, 2017.

\bibitem{heusel2017gans}
Martin Heusel, Hubert Ramsauer, Thomas Unterthiner, Bernhard Nessler, and Sepp
  Hochreiter.
\newblock Gans trained by a two time-scale update rule converge to a local nash
  equilibrium.
\newblock In {\em NeurIPS}, pages 6626--6637, 2017.

\bibitem{Jeong21}
Jongheon Jeong and Jinwoo Shin.
\newblock Training gans with stronger augmentations via contrastive
  discriminator.
\newblock In {\em ICLR}, 2021.

\bibitem{NEURIPS2020_f490c742}
Minguk Kang and Jaesik Park.
\newblock Contragan: Contrastive learning for conditional image generation.
\newblock In H. Larochelle, M. Ranzato, R. Hadsell, M.~F. Balcan, and H. Lin,
  editors, {\em Advances in Neural Information Processing Systems}, volume~33,
  pages 21357--21369. Curran Associates, Inc., 2020.

\bibitem{karras2017progressive}
Tero Karras, Timo Aila, Samuli Laine, and Jaakko Lehtinen.
\newblock Progressive growing of gans for improved quality, stability, and
  variation.
\newblock In {\em ICLR}, 2018.

\bibitem{karras2020training}
Tero Karras, Miika Aittala, Janne Hellsten, Samuli Laine, Jaakko Lehtinen, and
  Timo Aila.
\newblock Training generative adversarial networks with limited data.
\newblock {\em Advances in Neural Information Processing Systems},
  33:12104--12114, 2020.

\bibitem{karras2019style}
Tero Karras, Samuli Laine, and Timo Aila.
\newblock A style-based generator architecture for generative adversarial
  networks.
\newblock In {\em CVPR}, pages 4401--4410, 2019.

\bibitem{karras2020analyzing}
Tero Karras, Samuli Laine, Miika Aittala, Janne Hellsten, Jaakko Lehtinen, and
  Timo Aila.
\newblock Analyzing and improving the image quality of stylegan.
\newblock In {\em CVPR}, pages 8110--8119, 2020.

\bibitem{piq}
Sergey Kastryulin, Dzhamil Zakirov, and Denis Prokopenko.
\newblock {PyTorch Image Quality}: Metrics and measure for image quality
  assessment, 2019.
\newblock Open-source software available at
  https://github.com/photosynthesis-team/piq.

\bibitem{kingma2014adam}
Diederik Kingma and Jimmy Ba.
\newblock Adam: A method for stochastic optimization.
\newblock {\em ICLR}, 2014.

\bibitem{kynkaanniemi2019improved}
Tuomas Kynk{\"a}{\"a}nniemi, Tero Karras, Samuli Laine, Jaakko Lehtinen, and
  Timo Aila.
\newblock Improved precision and recall metric for assessing generative models.
\newblock In {\em NeurIPS}, 2019.

\bibitem{li2020few}
Yijun Li, Richard Zhang, Jingwan Lu, and Eli Shechtman.
\newblock Few-shot image generation with elastic weight consolidation.
\newblock In {\em NeurIPS}, 2020.

\bibitem{ZechunLiu19}
Zechun Liu, Haoyuan Mu, Xiangyu Zhang, Zichao Guo, Xin Yang, Kwang{-}Ting~(Tim)
  Cheng, and Jian Sun.
\newblock Metapruning: Meta learning for automatic neural network channel
  pruning.
\newblock In {\em ICCV}, pages 3296--3305, 2019.

\bibitem{mao2017least}
Xudong Mao, Qing Li, Haoran Xie, Raymond~YK Lau, Zhen Wang, and Stephen
  Paul~Smolley.
\newblock Least squares generative adversarial networks.
\newblock In {\em ICCV}, pages 2794--2802, 2017.

\bibitem{DBLP:journals/corr/abs-1801-04406}
Lars~M. Mescheder.
\newblock On the convergence properties of {GAN} training.
\newblock {\em CoRR}, abs/1801.04406, 2018.

\bibitem{Meyerson19}
Elliot Meyerson and Risto Miikkulainen.
\newblock Modular universal reparameterization: Deep multi-task learning across
  diverse domains.
\newblock In {\em NeurIPS}, 2019.

\bibitem{mo2020freeze}
Sangwoo Mo, Minsu Cho, and Jinwoo Shin.
\newblock Freeze the discriminator: a simple baseline for fine-tuning gans.
\newblock In {\em CVPR AI for Content Creation Workshop}, 2020.

\bibitem{naeem2020reliable}
Muhammad~Ferjad Naeem, Seong~Joon Oh, Youngjung Uh, Yunjey Choi, and Jaejun
  Yoo.
\newblock Reliable fidelity and diversity metrics for generative models.
\newblock In {\em ICML}, pages 7176--7185, 2020.

\bibitem{nilsback2008automated}
Maria-Elena Nilsback and Andrew Zisserman.
\newblock Automated flower classification over a large number of classes.
\newblock In {\em ICVGIP}, pages 722--729, 2008.

\bibitem{noguchi2019image}
Atsuhiro Noguchi and Tatsuya Harada.
\newblock Image generation from small datasets via batch statistics adaptation.
\newblock In {\em ICCV}, pages 2750--2758, 2019.

\bibitem{odena2017conditional}
Augustus Odena, Christopher Olah, and Jonathon Shlens.
\newblock Conditional image synthesis with auxiliary classifier gans.
\newblock In {\em ICML}, pages 2642--2651, 2017.

\bibitem{ojha2021few}
Utkarsh Ojha, Yijun Li, Jingwan Lu, Alexei~A Efros, Yong~Jae Lee, Eli
  Shechtman, and Richard Zhang.
\newblock Few-shot image generation via cross-domain correspondence.
\newblock In {\em CVPR}, pages 10743--10752, 2021.

\bibitem{DBLP:journals/corr/abs-1809-10889}
Zheyi Pan, Yuxuan Liang, Junbo Zhang, Xiuwen Yi, Yong Yu, and Yu Zheng.
\newblock Hyperst-net: Hypernetworks for spatio-temporal forecasting.
\newblock {\em CoRR}, abs/1809.10889, 2018.

\bibitem{perarnau2016invertible}
Guim Perarnau, Joost Van De~Weijer, Bogdan Raducanu, and Jose~M {\'A}lvarez.
\newblock Invertible conditional gans for image editing.
\newblock In {\em NeurIPS}, 2016.

\bibitem{Perez18}
Ethan Perez, Florian Strub, Harm de Vries, Vincent Dumoulin, and Aaron~C.
  Courville.
\newblock Film: Visual reasoning with a general conditioning layer.
\newblock In {\em AAAI}, pages 3942--3951, 2018.

\bibitem{Shahbazi_2021_CVPR}
Mohamad Shahbazi, Zhiwu Huang, Danda~Pani Paudel, Ajad Chhatkuli, and Luc
  Van~Gool.
\newblock Efficient conditional gan transfer with knowledge propagation across
  classes.
\newblock In {\em CVPR}, pages 12167--12176, 2021.

\bibitem{Shen18}
Falong Shen, Shuicheng Yan, and Gang Zeng.
\newblock Neural style transfer via meta networks.
\newblock In {\em CVPR}, pages 8061--8069, 2018.

\bibitem{shen2019interpreting}
Yujun Shen, Jinjin Gu, Xiaoou Tang, and Bolei Zhou.
\newblock Interpreting the latent space of gans for semantic face editing.
\newblock In {\em CVPR}, 2020.

\bibitem{shu2017neural}
Zhixin Shu, Ersin Yumer, Sunil Hadap, Kalyan Sunkavalli, Eli Shechtman, and
  Dimitris Samaras.
\newblock Neural face editing with intrinsic image disentangling.
\newblock In {\em CVPR}, pages 5541--5550, 2017.

\bibitem{Ivan21}
Ivan Skorokhodov, Savva Ignatyev, and Mohamed Elhoseiny.
\newblock Adversarial generation of continuous images.
\newblock In {\em CVPR}, pages 10753--10764, 2021.

\bibitem{Oswald20}
Johannes von Oswald, Christian Henning, Jo{\~{a}}o Sacramento, and Benjamin~F.
  Grewe.
\newblock Continual learning with hypernetworks.
\newblock In {\em ICLR}, 2020.

\bibitem{wang2020minegan}
Yaxing Wang, Abel Gonzalez-Garcia, David Berga, Luis Herranz, Fahad~Shahbaz
  Khan, and Joost van~de Weijer.
\newblock Minegan: effective knowledge transfer from gans to target domains
  with few images.
\newblock In {\em CVPR}, pages 9332--9341, 2020.

\bibitem{wang2021minegan++}
Yaxing Wang, Abel Gonzalez-Garcia, Chenshen Wu, Luis Herranz, Fahad~Shahbaz
  Khan, Shangling Jui, and Joost van~de Weijer.
\newblock Minegan++: Mining generative models for efficient knowledge transfer
  to limited data domains.
\newblock {\em arXiv preprint arXiv:2104.13742}, 2021.

\bibitem{wang2021transferi2i}
Yaxing Wang, H{\'e}ctor Laria, Joost van~de Weijer, Laura Lopez-Fuentes, and
  Bogdan Raducanu.
\newblock Transferi2i: Transfer learning for image-to-image translation from
  small datasets.
\newblock In {\em Proceedings of the IEEE/CVF International Conference on
  Computer Vision}, pages 14010--14019, 2021.

\bibitem{wang2018transferring}
Yaxing Wang, Chenshen Wu, Luis Herranz, Joost van~de Weijer, Abel
  Gonzalez-Garcia, and Bogdan Raducanu.
\newblock Transferring gans: generating images from limited data.
\newblock In {\em ECCV}, pages 218--234, 2018.

\bibitem{wang2020deepi2i}
Yaxing Wang, Lu Yu, and Joost van~de Weijer.
\newblock Deepi2i: Enabling deep hierarchical image-to-image translation by
  transferring from gans.
\newblock {\em NeurIPS}, 2020.

\bibitem{Zhirong18}
Zhirong Wu, Yuanjun Xiong, Stella~X. Yu, and Dahua Lin.
\newblock Unsupervised feature learning via non-parametric instance-level
  discrimination.
\newblock In {\em CVPR}, 2018.

\bibitem{yu2021dual}
Ning Yu, Guilin Liu, Aysegul Dundar, Andrew Tao, Bryan Catanzaro, Larry~S
  Davis, and Mario Fritz.
\newblock Dual contrastive loss and attention for gans.
\newblock In {\em Proceedings of the IEEE/CVF International Conference on
  Computer Vision}, pages 6731--6742, 2021.

\bibitem{Jure21}
Jure Zbontar, Li Jing, Ishan Misra, Yann LeCun, and St{\'{e}}phane Deny.
\newblock Barlow twins: Self-supervised learning via redundancy reduction.
\newblock In {\em ICML}, 2021.

\bibitem{ZhangRenUrtasun19}
Chris Zhang, Mengye Ren, and Raquel Urtasun.
\newblock Graph hypernetworks for neural architecture search.
\newblock In {\em ICLR}, 2019.

\bibitem{zhao2020leveraging}
Miaoyun Zhao, Yulai Cong, and Lawrence Carin.
\newblock On leveraging pretrained gans for limited-data generation.
\newblock {\em ICML}, 2020.

\bibitem{Shengyu20}
Shengyu Zhao, Zhijian Liu, Ji Lin, Jun{-}Yan Zhu, and Song Han.
\newblock Differentiable augmentation for data-efficient {GAN} training.
\newblock In {\em NeurIPS}, 2020.

\bibitem{zhou2017places}
Bolei Zhou, Agata Lapedriza, Aditya Khosla, Aude Oliva, and Antonio Torralba.
\newblock Places: A 10 million image database for scene recognition.
\newblock {\em TPAMI}, 40(6):1452--1464, 2017.

\bibitem{zhu2020domain}
Jiapeng Zhu, Yujun Shen, Deli Zhao, and Bolei Zhou.
\newblock In-domain gan inversion for real image editing.
\newblock {\em ECCV}, 2020.

\end{thebibliography}
}


\appendix
\clearpage

\begin{figure*}
    \centering
    \includegraphics[width=\linewidth]{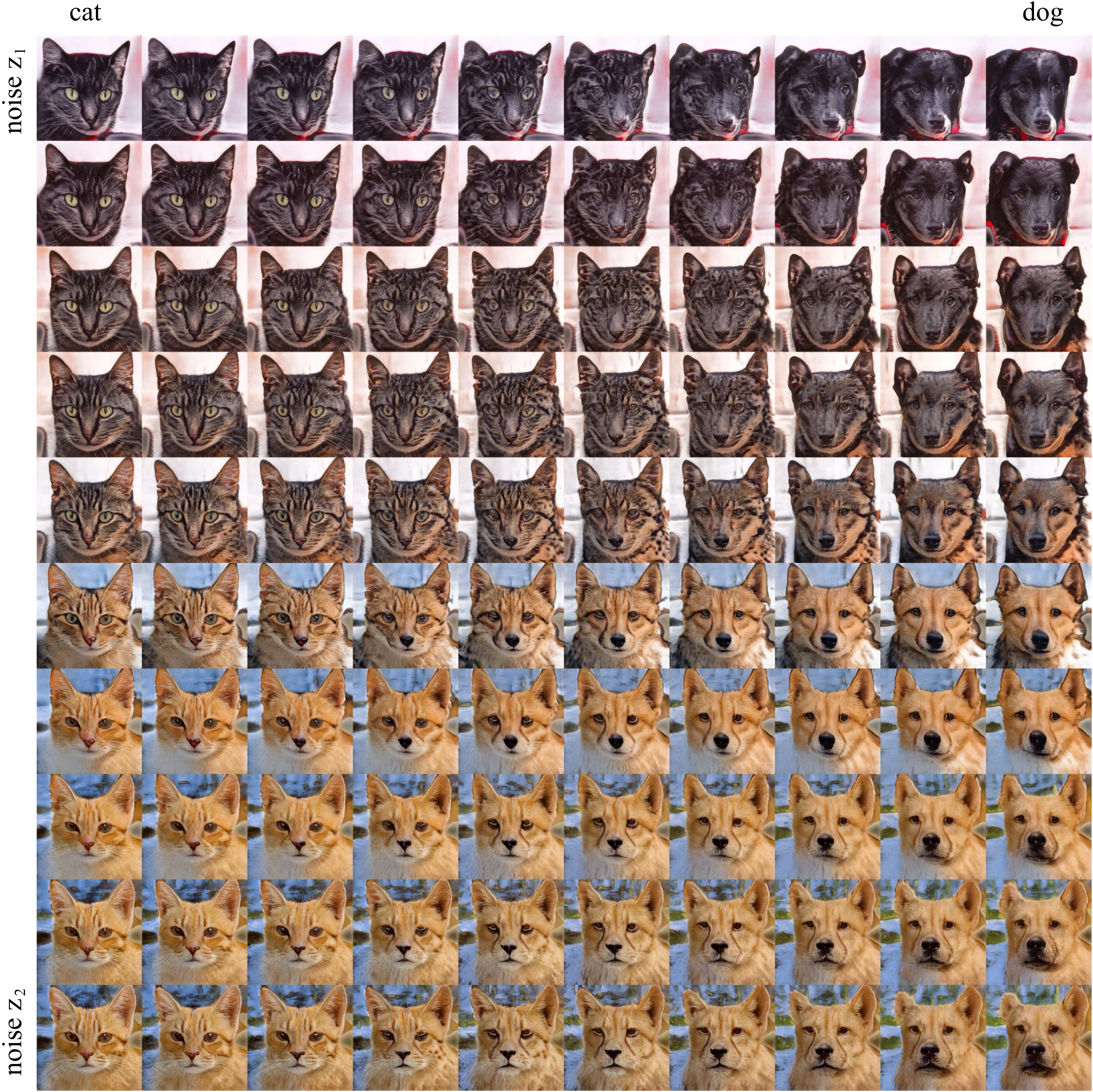}
    \caption{Sampling of class interpolation (left to right) versus noise interpolation (up to down). The hypernetwork has learned to keep every aspect of the style of an image intact, including background, while changing the class. The style mechanism was frozen since the beginning of the transfer learning training from human to animal faces.
    }
    \label{fig:suppl:interpolation_class_noise}
\end{figure*}

\begin{figure}
    \centering
    \includegraphics[width=\linewidth]{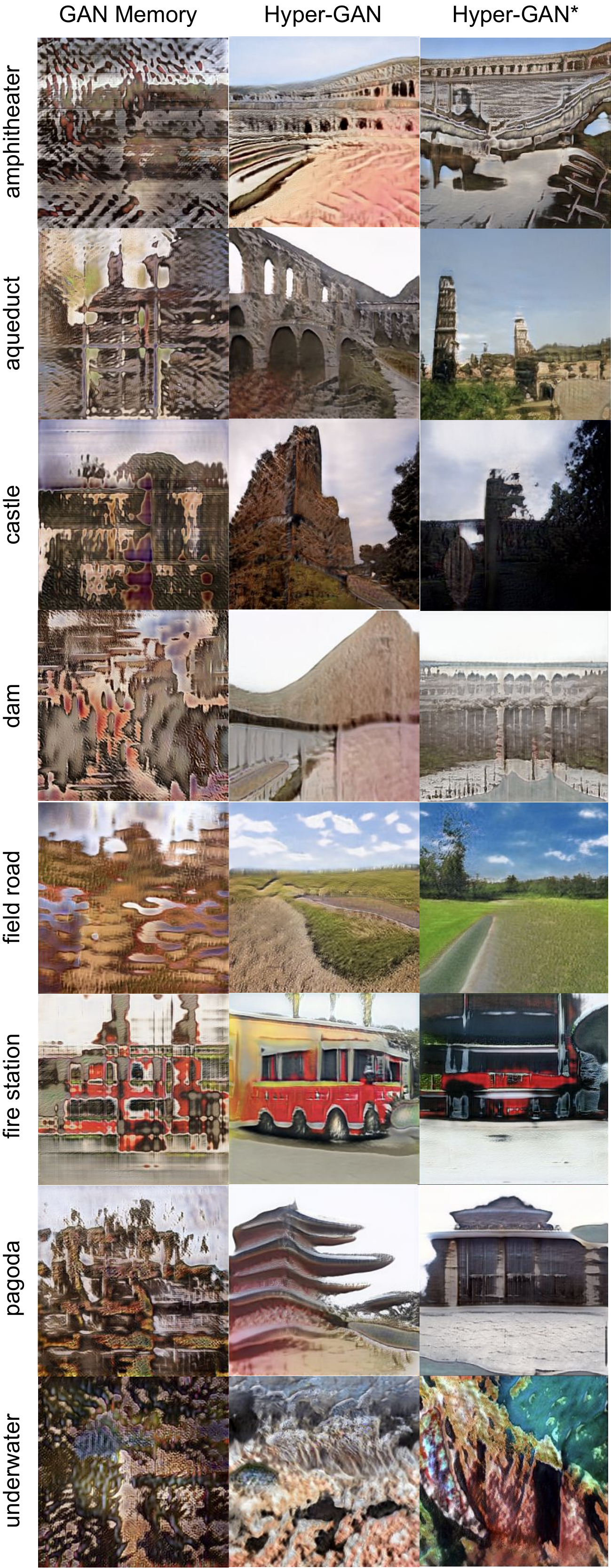}
    \caption{Transfer learning from Animal Faces (AFHQ) to a very distant domain (Places365).
    }
    \label{fig:suppl:places}
\end{figure}

\section{Discriminator details}
\begin{figure}[ht]
    \centering
    \includegraphics[width=0.95\linewidth]{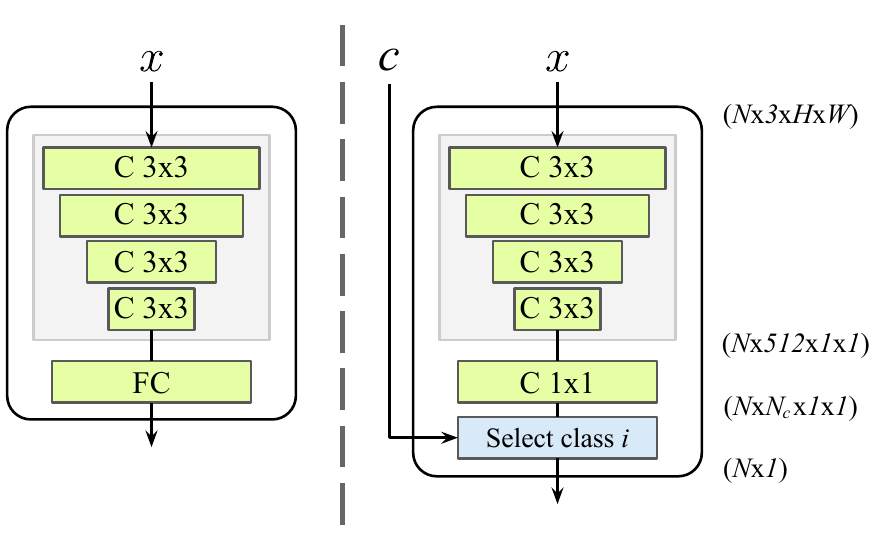}
    \caption{Original discriminator (left), modified discriminator (right). Batch dimensions at the right-most side for easier visualization. The final layer that comes from the convolutions is modified to output $N_c$ number of classes, and the correct one is picked to compute the final loss.
    }
    \label{fig:suppl:discriminator}
\end{figure}

A diagram of the discriminator and its modifications detailed in the experiments section can be seen in \Cref{fig:suppl:discriminator}. As underlined in the paper, it is worth noting that this module is finetuned from the source domain. Introducing weight modulation into it leads to training instability and worse generation quality overall.

\section{Additional baseline comments}
Results of mentioned baselines are shown in Table 4 for experiments on several domains. From these results, we can appreciate that \textit{cGANTransfer} performs badly when the number of previous learned classes is low, since its generation power comes precisely from combining these previous classes. For instance, the first column sets the problem to perform transfer learning from FFHQ (1 class) to AFHQ (3 classes), where this method seems to perform considerably worse than the proposed work, and also worse than simply learning the batch normalization statistics, as \textit{GAN Memory}.
When the number of source domain classes for the transfer is slightly higher, as AFHQ (3 classes) to CelebA-HQ (2 classes), the result is somewhat better since it is allowed more expressiveness, but still lacking quality depending on the closeness of the target domain.

Results for several metrics on transfer learning from pre-trained FFHQ model to AFHQ dataset are shown in Table 5. We can see how the proposed method performs better than simply learn the normalization statistics for each class as in \textit{GAN Memory} in terms of quality and diversity, since the knowledge of other classes learned concurrently can be propagated among all of them, resulting in improved time and data efficiency during training. We have already mentioned how \textit{cGANTransfer} quality degrades when not enough pre-trained classes are given to produce an interpolation. However, we can see how the diversity is better than its quality. We assume this occurs because it can produce close-enough interpolations to resemble the target class, but it doesn't have a meaningful basis (\ie a sufficient number of pre-trained classes) in order to form a combination of significant quality.

To perform experiments for cGANTransfer, a BigGAN model trained ImageNet was finetuned on FFHQ dataset. We used the checkpoint\footnote{\url{https://github.com/ajbrock/BigGAN-PyTorch}} of the highest resolution publicly available.
The model with the best FID before collapse was used as the base for this method. The same was performed for AFHQ dataset.

\section{Implementation details}
As the base of our method, we use a public StyleGAN implementation~\footnote{\url{https://github.com/rosinality/style-based-gan-pytorch}}, which while it is not official, it mostly reproduces results from the original paper. As already mentioned, we keep all hyperparameters from the original paper but fix the resolution growth to the final one, then apply all the methods explained.

In the original StyleGAN paper, it is mentioned training instability due to the depth of the mapping network. We experience a similar incident and therefore take the same solution of reducing the learning rate for the class network two orders of magnitude relative to the main network. 

For the evaluation metrics, we use a ready-made package \cite{piq} for \textit{FID}, \textit{KID} and \textit{Precision \& Recall}, which uses the original Inception feature extractor weights, ported to PyTorch. \textit{Density \& Coverage} metrics have been implemented as a package extension, also included in this paper. Perceptual path similarity implementation is taken from \footnote{\url{https://github.com/rosinality/stylegan2-pytorch/blob/master/ppl.py}} applying default center crop.

\section{Architecture specifics}
In this paper, we propose a novel transfer learning strategy from unconditional GAN to conditional GAN by introducing hypernetwork-based
adaptive weight modulation. Here we will detail the concrete architecture we used, and the changes applied to it.

\begin{figure*}
    \centering
      \includegraphics[width=0.7\linewidth]{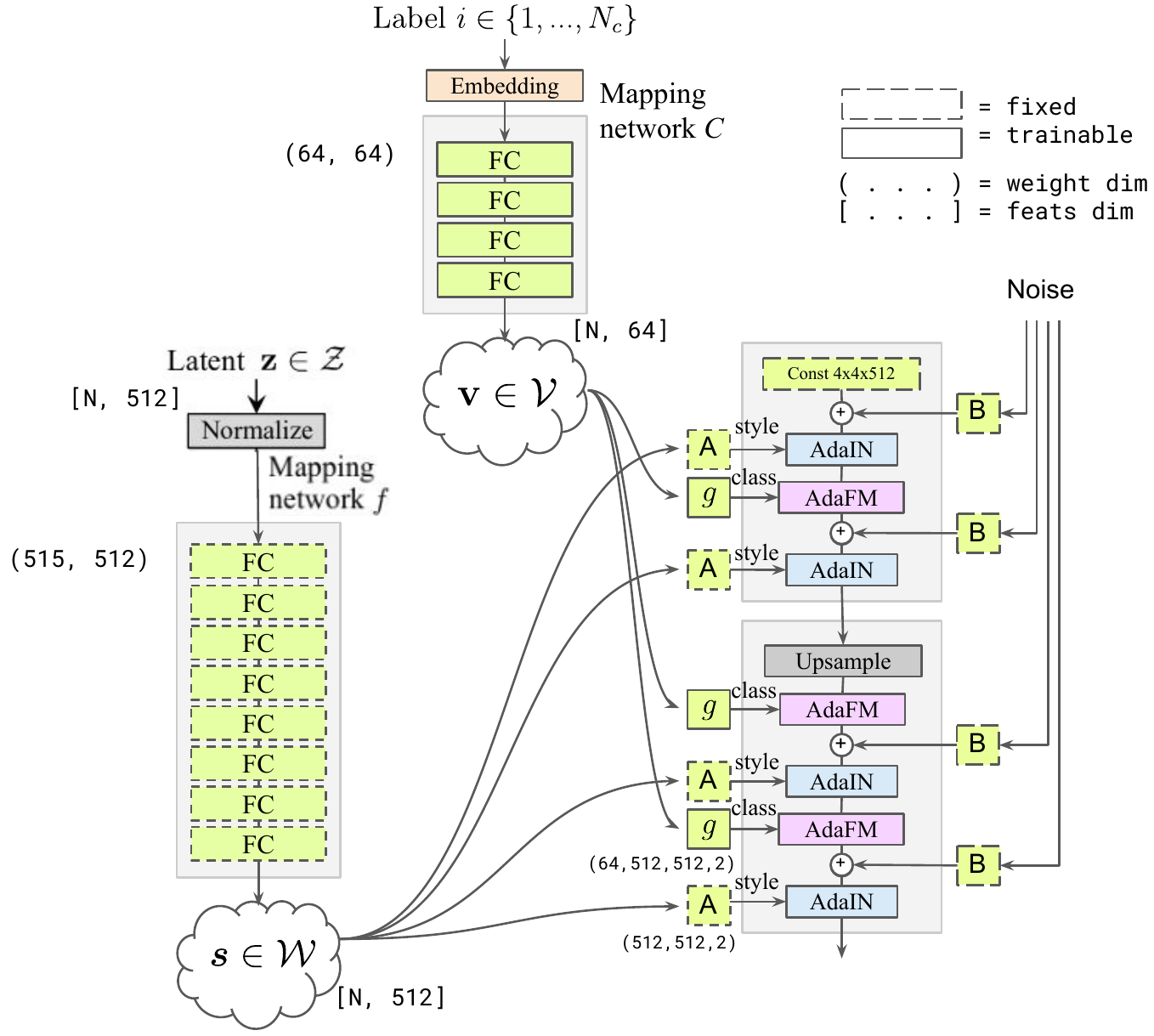}
    \caption{Hypernetwork architecture for transfer learning from unconditional GAN to conditional GAN, 
    which wrap around the pre-trained weights (\ie, the convolutional  and fully connected layers). The network \textit{C} takes as input the class index \textit{i}, and outputs the class embedding $\mathbf{v}$, which is fed into the hypernetwork generators \textit{g}. 
    Adapted from \cite{karras2019style}.
    }
    \label{fig:suppl:general}
\end{figure*}

\Cref{fig:suppl:general} shows the changes made to a vanilla StyleGAN~\cite{karras2019style}. The style branch is frozen since we want to keep the learned transformations (pose rotations, color changes, etc.) from the source domain, \ie \textit{FFHQ}, unchanged. We do not see a loss in performance when transferred to other datasets (see \cref{fig:suppl:interpolation_class_noise}). The class network \textit{C} is very similar to the original mapping network and also generates an embedding space, in this case $\mathcal{V}$, for classes, with the difference that the input comes from a learned class-embedding. The information then comes into each convolution layer to modulate the weights, as explained in the main paper.

\section{Domain information injection} \label{sec:suppl:domain_mod_method}
\begin{figure}
    \centering
    \begin{subfigure}[b]{0.45\linewidth}
      \includegraphics[width=\linewidth]{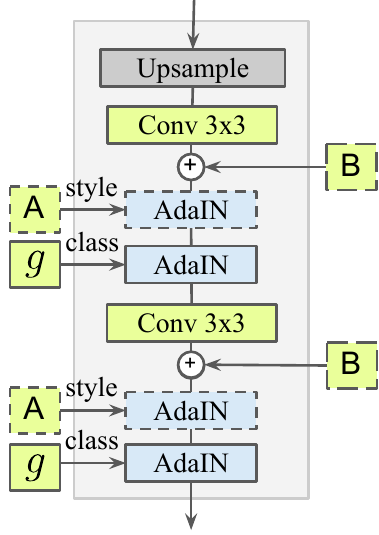}
      \caption{Naive class introduction.}
      \label{fig:suppl:adain-adain-1}
    \end{subfigure}
    \hfill
    \begin{subfigure}[b]{0.45\linewidth}
      \includegraphics[width=\linewidth]{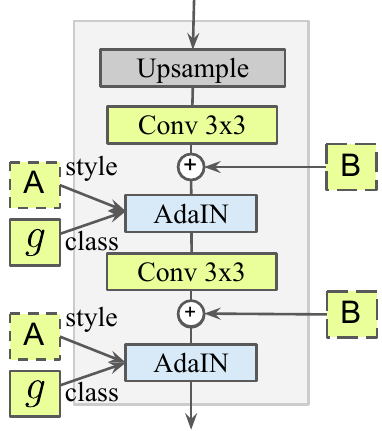}
      \caption{Simplified.}
      \label{fig:suppl:adain-adain-2}
    \end{subfigure}
    \caption{Different fusion strategies. Dashed boxes are frozen during training.}
    \label{fig:suppl:adain-adain}
\end{figure}

We specify here a different class information introduction technique to the one in the main paper.
Since the normalization from one block will destroy the information included by the other (\cref{fig:suppl:adain-adain-1}), we can fix this and simplify the formulation by combining style and class as $\boldsymbol{\gamma}_{(\mathbf{s}, \mathbf{v})}, \boldsymbol{\beta}_{(\mathbf{s}, \mathbf{v})} = g(\mathbf{s}) + g(\mathbf{v};\Phi)$, corresponding to Eq. (3), as seen in \cref{fig:suppl:adain-adain-2}.
Nevertheless, we have experienced consistent underperformance when compared to the current weight modulation technique used.


\begin{figure*}
    \centering
    \includegraphics[width=\linewidth]{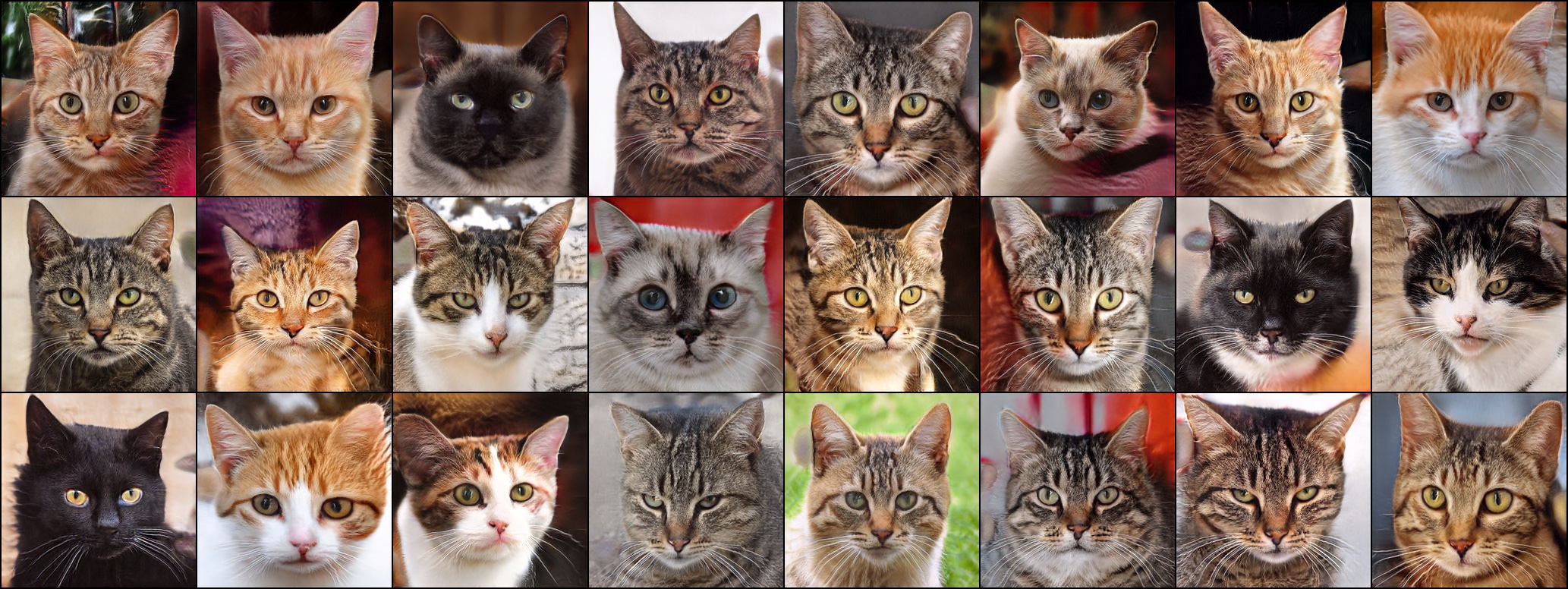}
    \includegraphics[width=\linewidth]{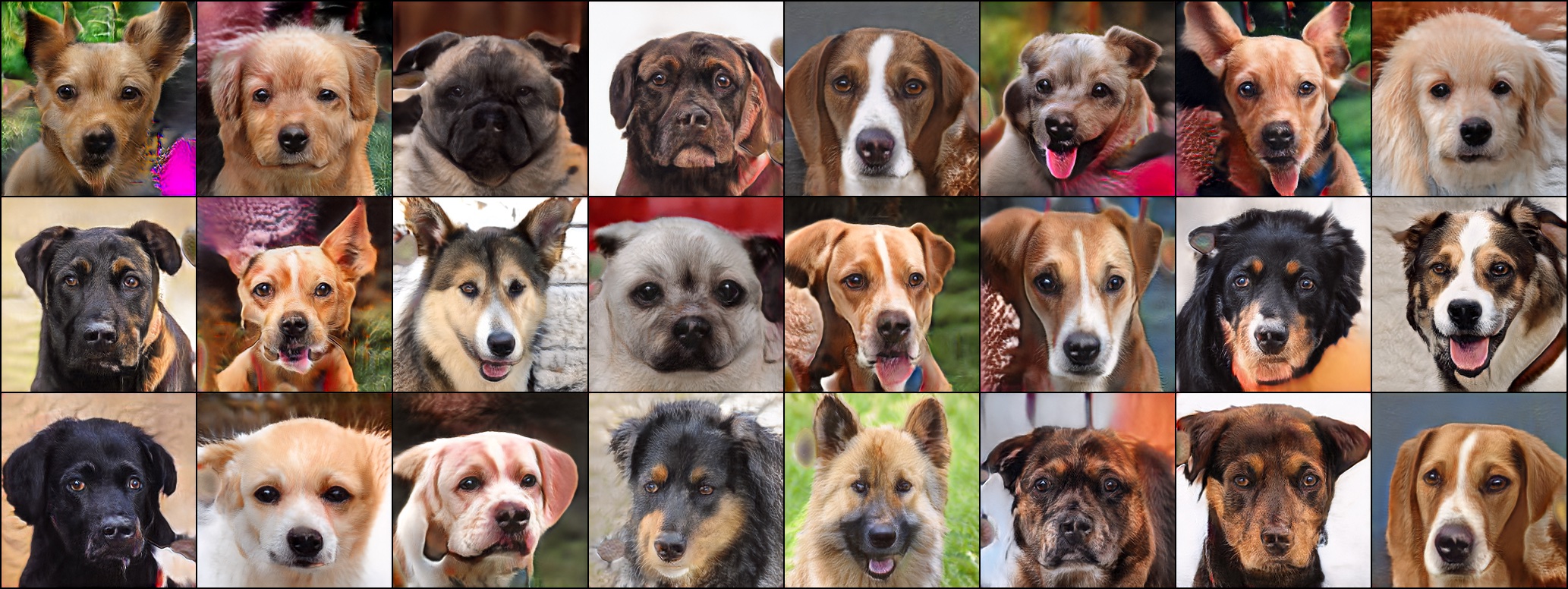}
    \includegraphics[width=\linewidth]{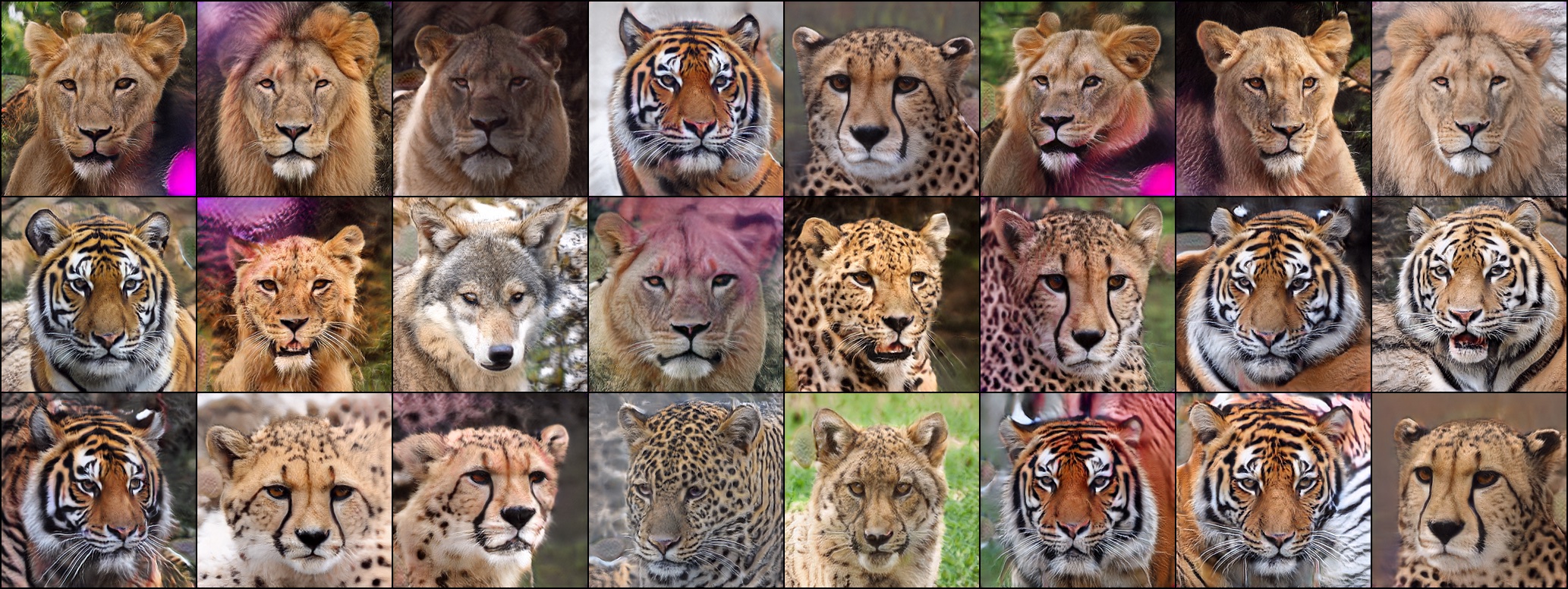}
    \caption{Unfiltered conditional generations. Three rows of cats, dogs and wildlife respectively. The same noise is applied among classes (poses, background, etc).
    \label{fig:suppl:unfiltered-gen}
    }
\end{figure*}

\begin{figure*}
    \centering
    \includegraphics[width=\linewidth]{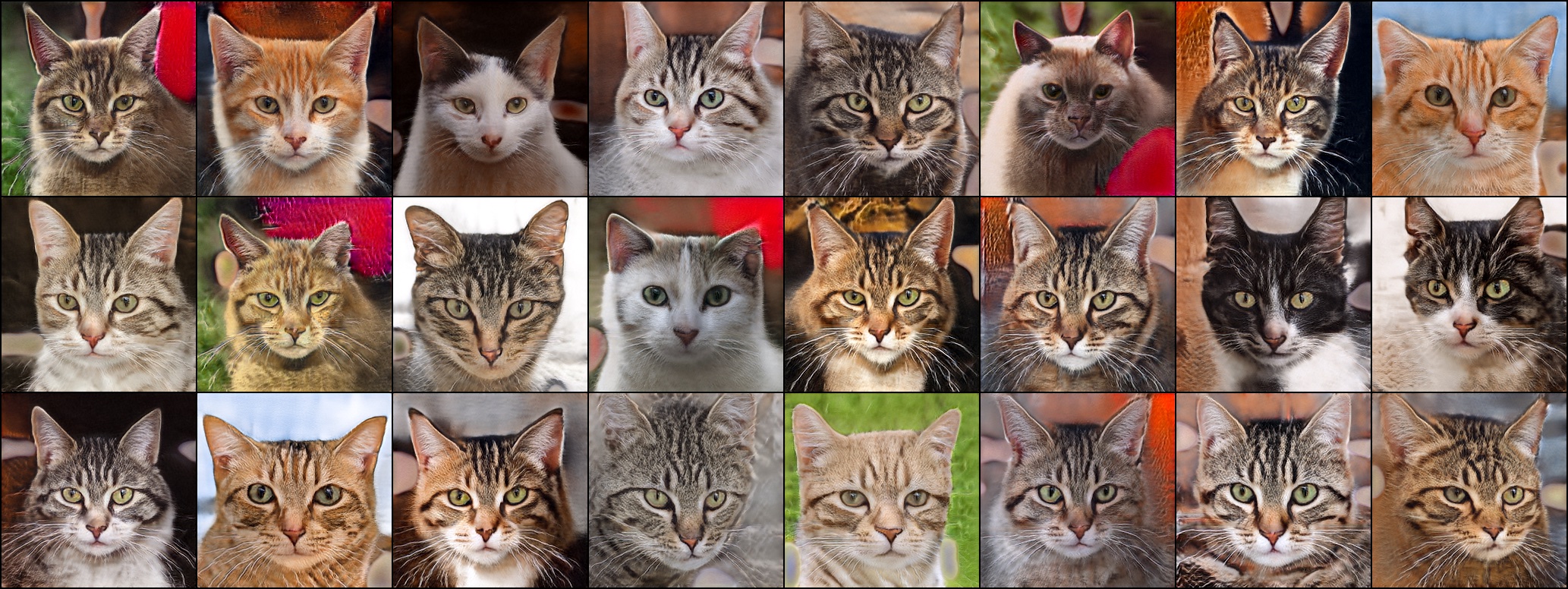}
    \includegraphics[width=\linewidth]{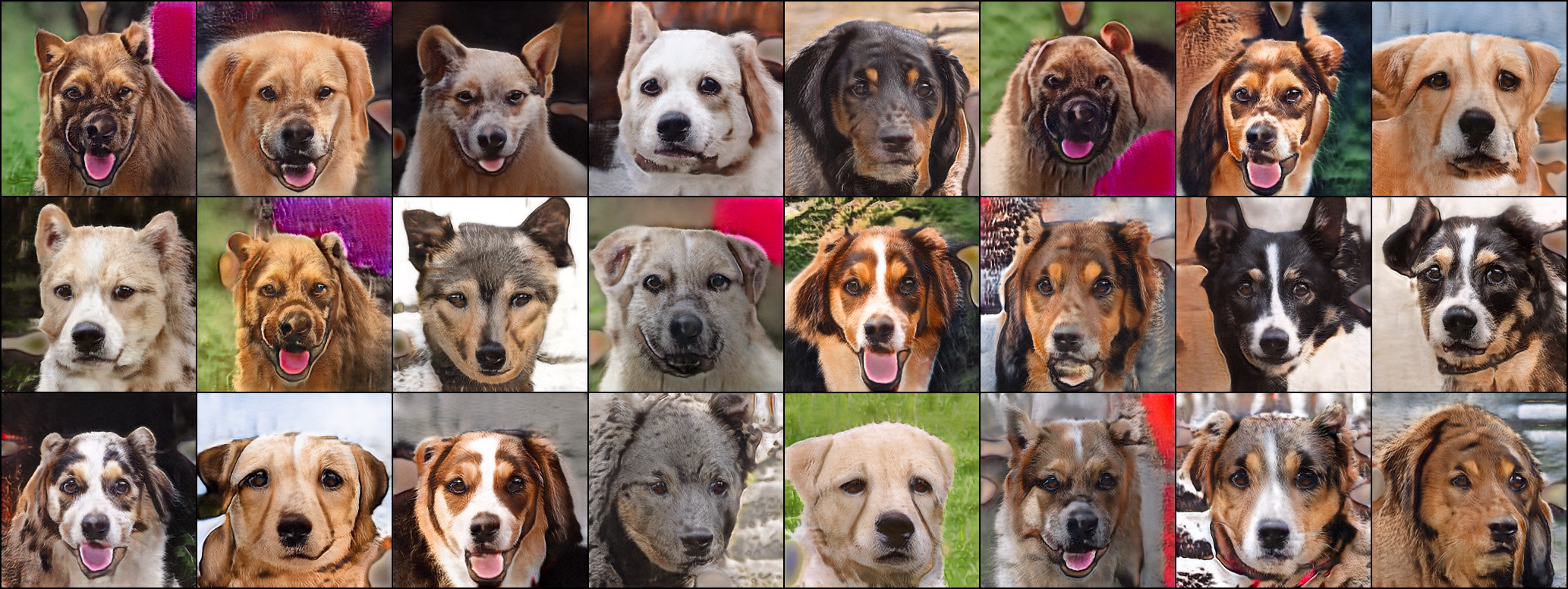}
    \includegraphics[width=\linewidth]{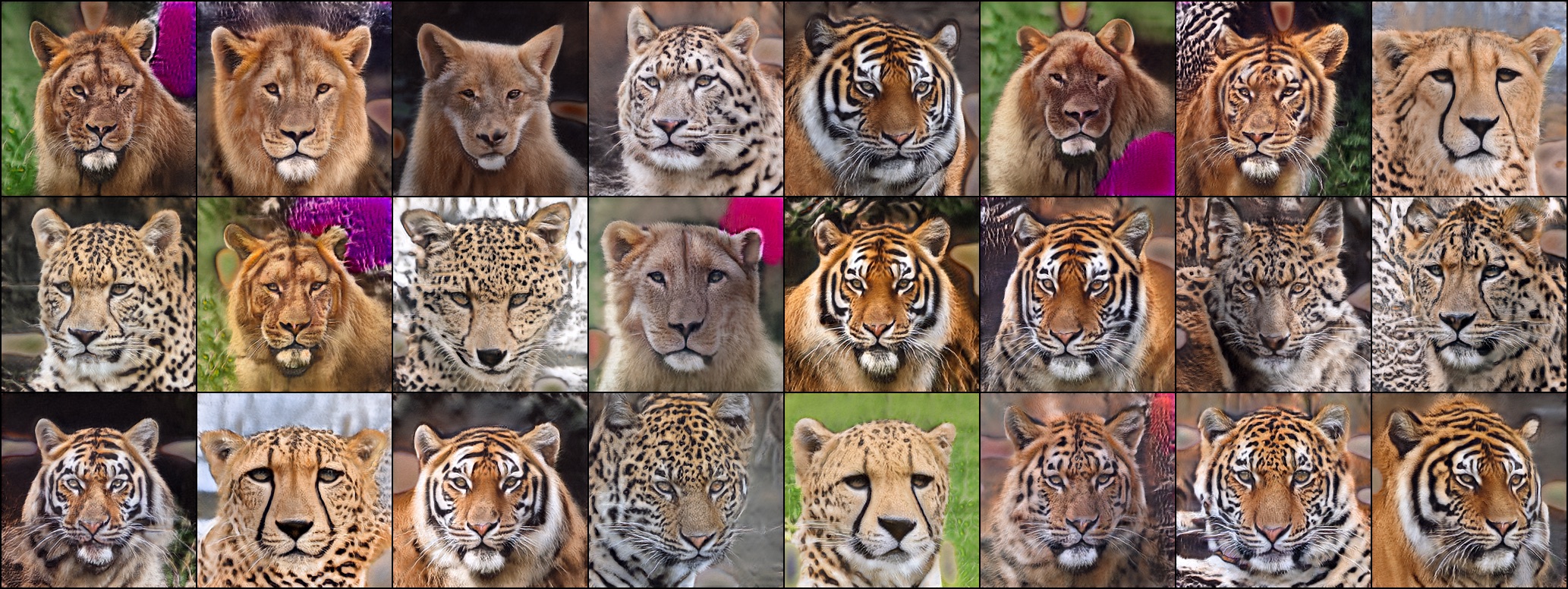}
    \caption{Unfiltered conditional generations (config.\ {\small B}, worse than \cref{fig:suppl:unfiltered-gen}). Three rows of cats, dogs and wildlife respectively. The same noise is applied among classes (poses, background, etc.).
    }
\end{figure*}

\begin{figure*}
    \centering
    \includegraphics[width=\linewidth]{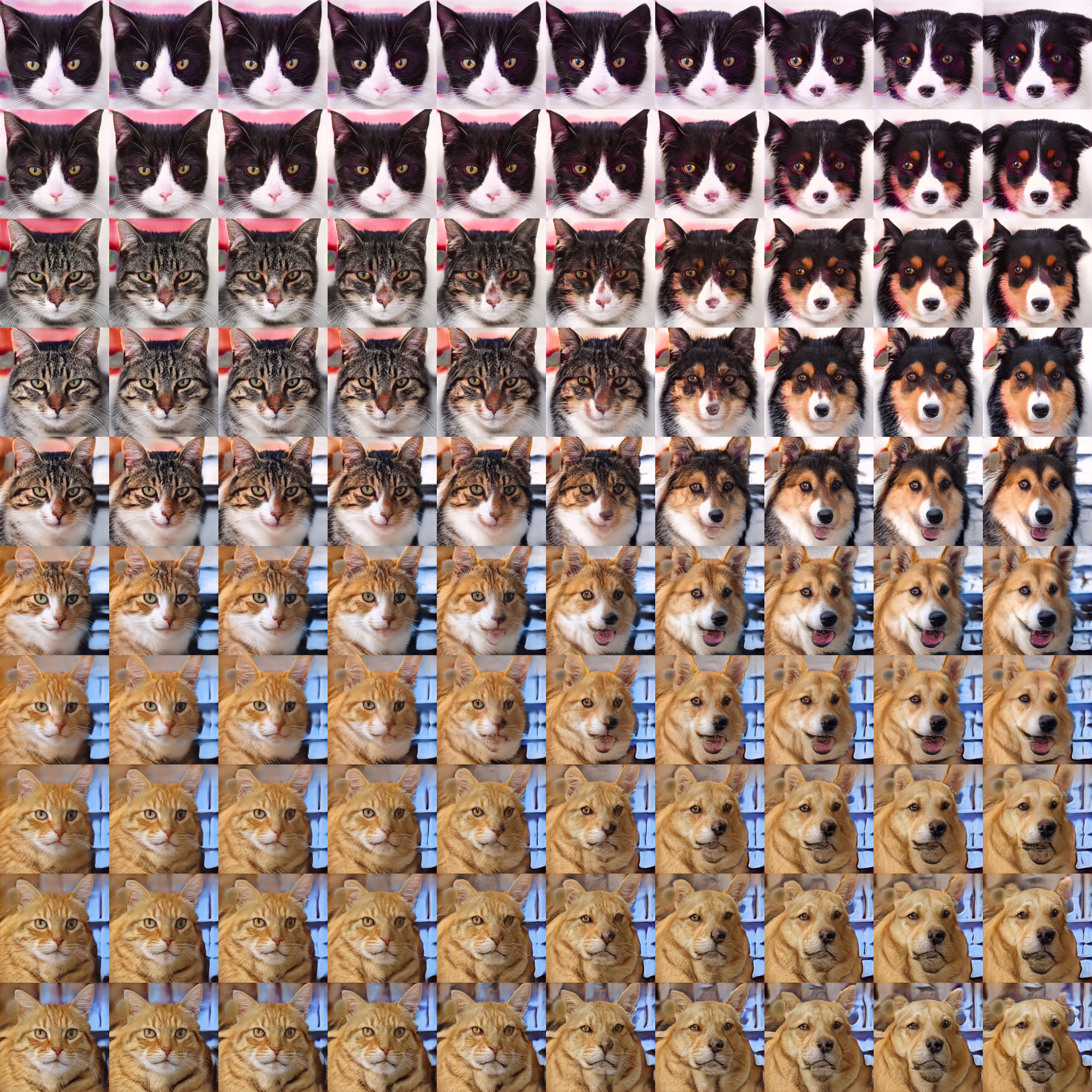}
    \caption{Sampling of class interpolation (left to right) versus noise interpolation (up to down).
    }
\end{figure*}

\begin{figure*}
    \centering
    \includegraphics[width=\linewidth]{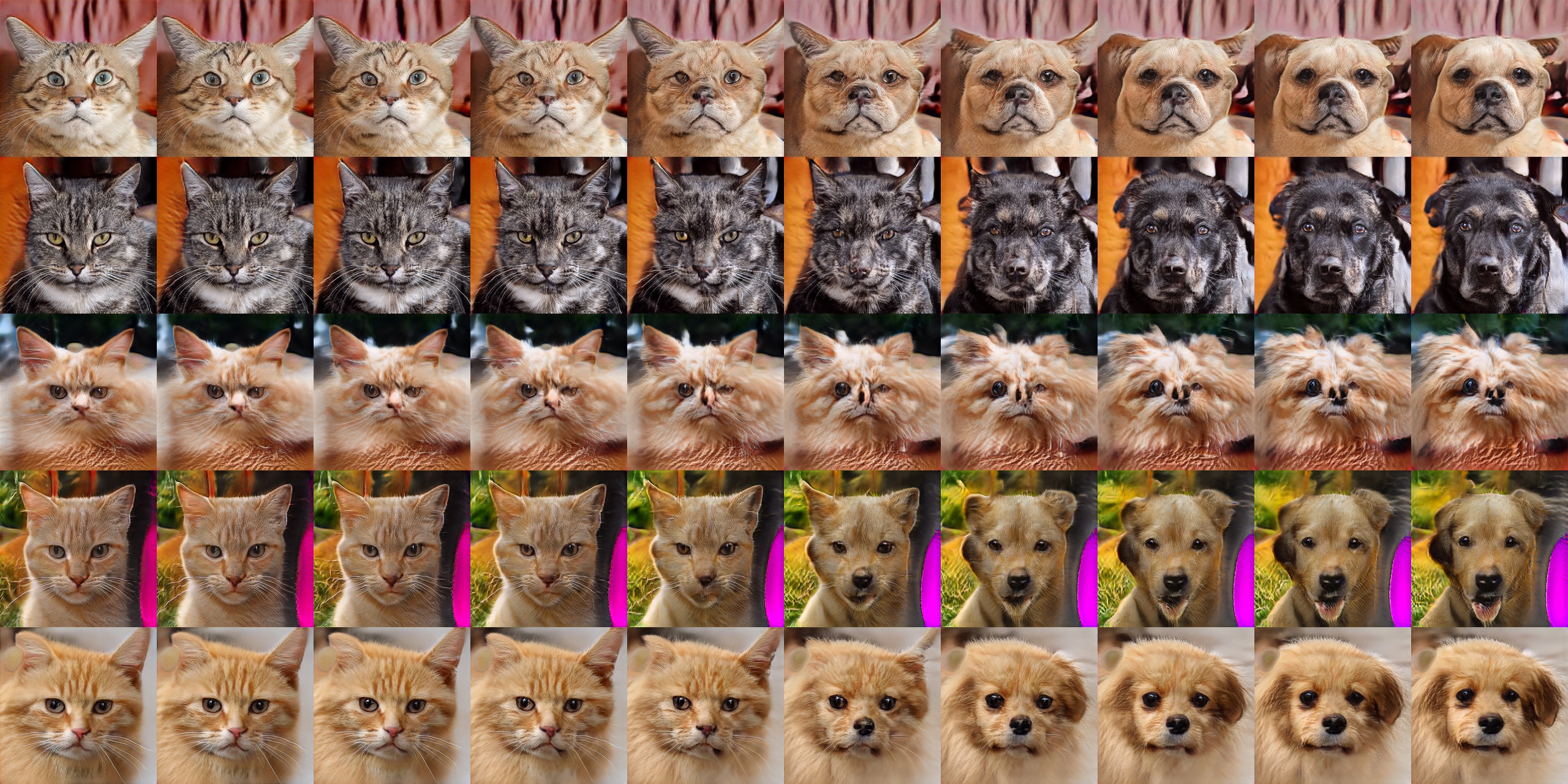}
    \caption{Sampling of class interpolation between cat and dog.
    }
\end{figure*}

\begin{figure*}
    \centering
    \includegraphics[width=\linewidth]{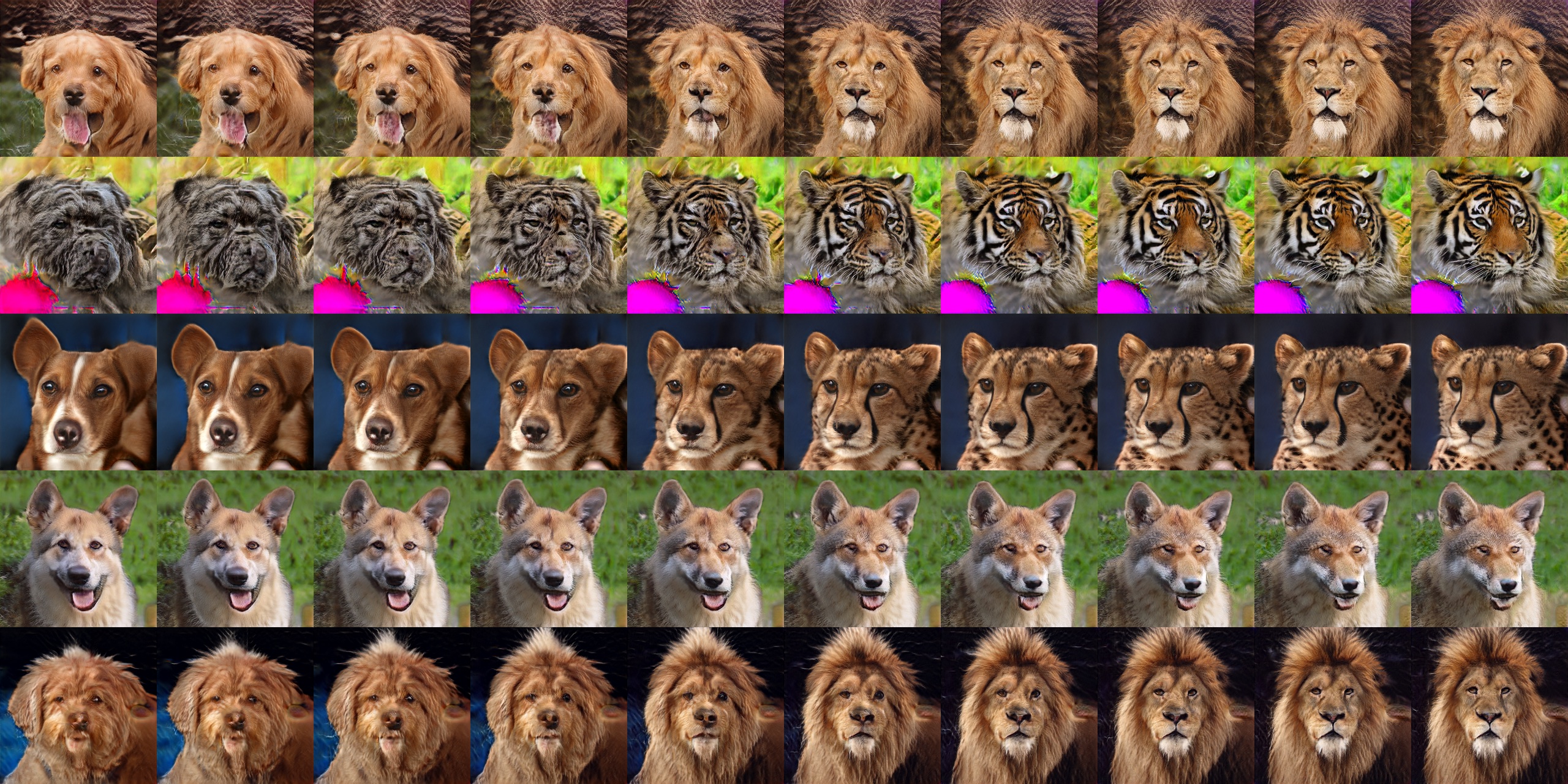}
    \caption{Sampling of class interpolation between dog and wild life.
    }
\end{figure*}

\begin{figure*}
    \centering
    \includegraphics[width=\linewidth]{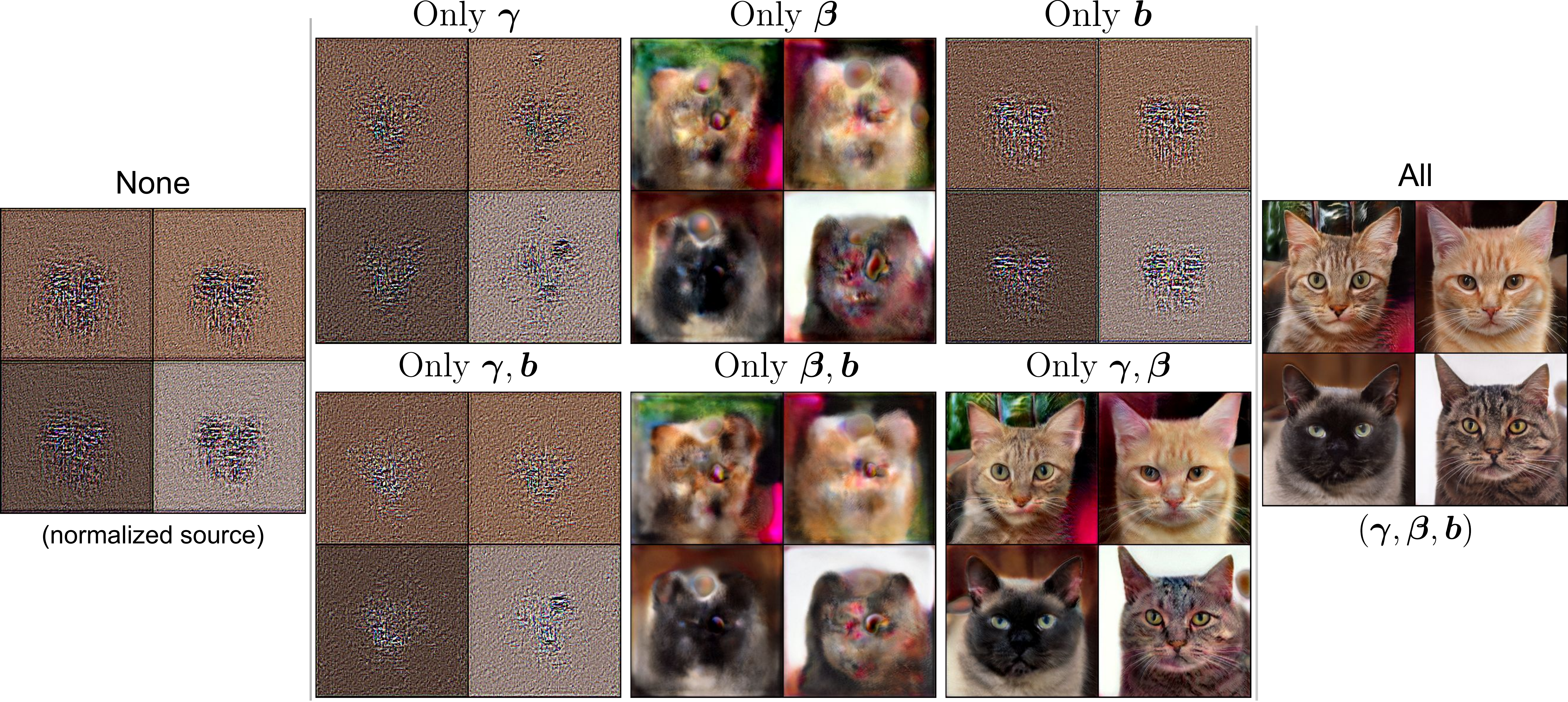}
    \caption{Effect of the modulation parameters on the domain transfer.
    }
    \label{fig:suppl:mod-params-big}
\end{figure*}

\end{document}